\documentclass[preprint,12pt]{elsarticle}

\usepackage{amssymb}
\usepackage{amsmath}
\usepackage{amsthm}

\usepackage{booktabs}    
\usepackage{multirow}    
\usepackage[colorlinks=true,linkcolor=blue,citecolor=blue,urlcolor=blue]{hyperref}    
\usepackage{url}         
\usepackage{algorithm}   
\usepackage{algorithmic} 
\usepackage{xcolor}      

\usepackage{mathtools}      
\usepackage{bm}             
\usepackage{bbm}            
\usepackage{enumitem}       
\usepackage{cleveref}       
\usepackage{makecell} 

\usepackage{tikz}
\usetikzlibrary{patterns}
\usepackage{pgfplots}
\pgfplotsset{compat=1.18}
\usepackage{subcaption}  

\usepackage{lineno}
\usepackage{amsmath}
\usepackage{fontawesome5}
\usepackage{hyperref}
\usepackage{fancyhdr}
\usepackage{lipsum}

\DeclareMathOperator*{\argmin}{arg\,min}

\DeclareMathOperator*{\texttext}{\text{text}}
\DeclareMathOperator*{\textimage}{\text{text--image}}
\DeclareMathOperator*{\imagetext}{\text{image--text}}
\DeclareMathOperator*{\imageimage}{\text{image}}

\newcommand{\crossmodaldataset}{\textsc{\small XM3600}}

\graphicspath{{figures/}}

\journal{Expert Systems with Applications}

\begin{document}

\begin{frontmatter}



\title{ViCLIP-OT: The First Foundation Vision-Language Model for Vietnamese Image--Text Retrieval with Optimal Transport}

\author[ctu]{Quoc-Khang Tran}
\ead{tqkhang@ctu.edu.vn}

\author[ctu]{Minh-Thien Nguyen}

\author[ctu]{Nguyen-Khang Pham\corref{cor1}}
\cortext[cor1]{Corresponding author.}

\affiliation[ctu]{
  organization={Can Tho University},
  city={Can Tho},
  country={Vietnam}
}







\begin{abstract}
Image-text retrieval has become a fundamental component in intelligent multimedia systems; however, most existing vision--language models are optimized for high-resource languages and remain suboptimal for low-resource settings such as Vietnamese. This work introduces ViCLIP-OT, a foundation vision--language model specifically designed for Vietnamese image--text retrieval. The proposed framework integrates CLIP-style contrastive learning with a Similarity-Graph Regularized Optimal Transport (SIGROT) loss to enhance global cross-modal consistency and mitigate modality gap issues. Extensive experiments on three Vietnamese benchmarks (UIT-OpenViIC, KTVIC, and Crossmodal-3600) demonstrate that ViCLIP-OT consistently outperforms CLIP and SigLIP baselines in both in-domain and zero-shot settings. On UIT-OpenViIC, the model achieves an average Recall@K of 67.34\%, improving upon CLIP by 5.75 percentage points. In zero-shot evaluation on Crossmodal-3600, ViCLIP-OT surpasses CLIP by 11.72 percentage points. Embedding-space analysis further confirms improved alignment and reduced modality gap. The results indicate that integrating SIGROT provides an effective and scalable strategy for cross-modal retrieval in low-resource languages, offering practical implications for intelligent multimedia retrieval systems in Vietnamese and other underrepresented linguistic contexts.

\end{abstract}












\end{frontmatter}




\fancyhf{}
\fancyhead[R]{\textit{Preprint submitted to Machine Vision and Applications}}
\fancyfoot[C]{\thepage}
\renewcommand\headrulewidth{0pt}
\pagestyle{fancy}

\section{Introduction}

Image--text retrieval~\cite{cao2022imagetextretrievalsurveyrecent} has achieved significant progress thanks to large-scale vision–language pre-training. Models such as CLIP~\cite{pmlr-v139-radford21a} and ALIGN~\cite{jia2021scaling} use dual-encoder architectures trained on hundreds of millions to billions of image--text pairs, allowing them to learn general multimodal representations and achieve strong retrieval performance.

However, most of these advances focus on high-resource languages, especially English. For low-resource languages like Vietnamese, the lack of large-scale image--text datasets and pre-trained models remains a major challenge. Until recently, only a limited number of public image-caption datasets were available for Vietnamese~\cite{10.1007/978-3-030-63007-2_57,pham2024ktvicvietnameseimagecaptioning,BUI2026117430,thapliyal2022crossmodal}, which limits the direct application of CLIP-style training. A common solution is to translate Vietnamese captions into English and apply English-based models, but this approach may introduce translation noise and cannot fully preserve language-specific meanings.

This work addresses the aforementioned gap by proposing ViCLIP-OT, a contrastive vision--language model tailored for Vietnamese image--text retrieval. ViCLIP-OT extends the CLIP dual-encoder architecture in two key aspects. First, it adopts strong and recent backbone models for each modality: a DINOv3-based~\cite{simeoni2025dinov3} vision transformer for image encoding, and a Vietnamese Sentence-BERT model pretrained on large-scale Vietnamese corpora for text encoding. Both encoders project images and texts into a shared embedding space and are jointly optimized using a contrastive loss, following the standard CLIP training paradigm. Second, ViCLIP-OT incorporates a Similarity-Graph Regularized Optimal Transport (SIGROT) loss based on OT~\cite{ambrosio2002optimaltransportmapsmongekantorovich,cuturi2013sinkhorndistanceslightspeedcomputation,montesuma2024recent,khamis24OT} to align image and text representations while preserving global relational structure among samples. A precomputed similarity graph encodes relationships among samples within each training batch, and an OT solver is employed to find a globally consistent cross-modal matching that respects such relational structure. This structure-aware alignment complements the instance-level contrastive objective, enabling the model to capture not only pairwise correspondences but also distributional relationships across modalities within each batch.

ViCLIP-OT is trained and evaluated on three Vietnamese image--text datasets that differ in domain and scale. The primary evaluation is conducted on UIT-OpenViIC~\cite{BUI2026117430}, a large-scale open-domain Vietnamese image captioning dataset comprising 13,100 images and 61,241 captions, featuring complex real-world scenes and diverse semantic content. This dataset is used to assess the model's overall retrieval performance under realistic and challenging conditions. In addition, zero-shot evaluation is performed on two held-out datasets to assess the robustness and transferability of the learned representations: KTVIC~\cite{pham2024ktvicvietnameseimagecaptioning}, a benchmark comprising 4,327 images and 21,635 captions depicting daily-life scenarios in Vietnam, and Crossmodal-3600~\cite{thapliyal2022crossmodal}, a geographically diverse collection of 3,600 photos tagged with human-generated reference captions in 36 languages. Since no fine-tuning is performed on either dataset, these evaluations directly measure how well the learned vision--language representations generalize across different domains.



\paragraph{Overview of Contributions}
The main contributions of this work are summarized as follows:
\begin{itemize}
    \item ViCLIP-OT is introduced as a foundation vision--language model for Vietnamese image--text retrieval, extending CLIP-style contrastive learning with an OT-based mechanism to capture fine-grained cross-modal alignments.

    \item SIGROT loss is proposed to enhance cross-modal alignment by leveraging relational structures among samples within each training batch, thereby improving representation consistency and mitigating modality gap.

    \item Extensive experiments on three Vietnamese image--text retrieval benchmarks demonstrate that ViCLIP-OT achieves state-of-the-art performance, with strong zero-shot generalization capability across datasets.
\end{itemize}

To the best of our knowledge, ViCLIP-OT is the first foundation vision--language model developed for Vietnamese at this scale, providing strong performance in image--text retrieval and cross-modal understanding. The pretrained models are publicly available to support reproducibility and future research.\footnote{\url{https://huggingface.co/collections/minhnguyent546/viclip-ot}}

\paragraph{Paper Organization}
The remainder of this paper is organized as follows. Section~\ref{sec:background} reviews related work on contrastive learning and optimal transport (OT). Section~\ref{sec:preliminaries} introduces preliminaries on image--text retrieval and OT. Section~\ref{sec:proposed-method} describes the ViCLIP-OT architecture and the formulation of the SIGROT loss. Section~\ref{sec:experiments} presents the datasets, evaluation metrics, and implementation details. Section~\ref{sec:results} reports and discusses the experimental results. Section~\ref{sec:ablation_study} provides ablation studies analyzing key components of the proposed method. Finally, Section~\ref{sec:conclusion} concludes the paper and outlines directions for future work.

\section{Background}\label{sec:background}

Contrastive learning~\cite{hu2024comprehensive} is a representation learning paradigm that aims to learn discriminative embeddings by pulling semantically related samples closer while pushing unrelated ones apart in a shared latent space~\cite{chen2020simpleframeworkcontrastivelearning}. In cross-modal tasks such as image--text retrieval, contrastive objectives align visual and textual representations by treating matched image--caption pairs as positives and mismatched pairs as negatives, enabling robust zero-shot transfer and effective cross-modal similarity estimation~\cite{pmlr-v139-radford21a,YANG2024110273,csizmadia2025distillclipdclipenhancing,schall2024optimizingclipmodelsimage}. While contrastive learning enforces instance-level alignment, Optimal Transport (OT) provides a complementary, distribution-level perspective by computing a cost-efficient transport plan between probability distributions, as originally formulated by Monge and Kantorovich~\cite{ambrosio2002optimaltransportmapsmongekantorovich}. In machine learning, OT has been widely applied to distribution alignment problems, including cross-modal representation learning~\cite{khamis24OT,montesuma2024recent}. Recent studies have shown that entropic regularized OT can bridge contrastive learning with distribution alignment, yielding more structured and semantically consistent embedding spaces~\cite{shi2023understanding,10.3389/fcomp.2024.1473457}. Together, contrastive learning and OT offer complementary mechanisms for cross-modal learning, where contrastive objectives capture pairwise semantic relationships and OT enhances global alignment across modalities.

\section{Preliminaries}\label{sec:preliminaries}





\subsection{Image-Text Retrieval}\label{subsec:preli-image-text-retrieval}

Image--text retrieval involves searching for the most relevant images given a text query (text-to-image retrieval) or finding the most relevant captions for a given image (image-to-text retrieval)~\cite{karpathy2015deep,chen2020uniter,pmlr-v139-radford21a}. Formally, let $\mathcal{D} = \{(x_i, t_i)\}_{i=1}^M$ denote a dataset of $M$ image--text pairs, where $x_i$ is an image and $t_i$ is its corresponding caption. The goal of image--text retrieval is to learn two encoders, $f_{\text{image}}$ and $f_{\text{text}}$, that map images and texts into a shared embedding space:
\begin{equation}
\mathbf{z}^{\text{image}}_i = f_{\text{image}}(x_i), \qquad
\mathbf{z}^{\text{text}}_i = f_{\text{text}}(t_i),
\end{equation}
where $\mathbf{z}^{\text{image}}_i, \mathbf{z}^{\text{text}}_i \in \mathbb{R}^d$ are $d$-dimensional embeddings. The similarity between an image and a text is typically measured using cosine similarity:
\begin{equation}
\text{sim}(x_i, t_j) = \frac{\mathbf{z}^{\text{image}}_i \cdot \mathbf{z}^{\text{text}}_j}{\|\mathbf{z}^{\text{image}}_i\|_2 \|\mathbf{z}^{\text{text}}_j\|_2}.
\end{equation}
During retrieval, given a query (image or text), the model ranks all candidates in the database based on their similarity scores and returns the top-$K$ most relevant results. The performance of image--text retrieval models is commonly evaluated using metrics such as Recall@K (R@K), which measures the proportion of queries for which the correct match is found within the top-$K$ retrieved results.

\subsection{Optimal Transport}\label{subsec:preli-optimal-transport}
Optimal Transport (OT) is a mathematical framework that seeks the most efficient way to transport mass from one probability distribution to another while minimizing a total transportation cost~\cite{ambrosio2002optimaltransportmapsmongekantorovich}. It provides a meaningful geometric distance between probability distributions, making it particularly useful for aligning data across different domains or modalities.

Formally, let $\mu \in \mathbb{R}^n_+$ and $\nu \in \mathbb{R}^m_+$ be two discrete probability measures such that $\sum_{i=1}^n \mu_i = \sum_{j=1}^m \nu_j = 1$. Given a cost matrix $C \in \mathbb{R}^{n \times m}_{+}$, where $C_{ij}$ represents the cost of moving a unit of mass from the $i$-th element of $\mu$ to the $j$-th element of $\nu$, the goal of OT is to find a transport plan $\gamma \in \mathbb{R}^{n \times m}_{+}$ that minimizes the total cost:
\begin{equation}
    \gamma^* = \argmin_{\gamma \in \Pi(\mu, \nu)} \langle \gamma, C \rangle_F = \argmin_{\gamma \in \Pi(\mu, \nu)} \sum_{i=1}^n \sum_{j=1}^m \gamma_{ij} C_{ij},
    \label{eq:ot_primal}
\end{equation}
where $\langle \cdot, \cdot \rangle_F$ denotes the Frobenius inner product, and $\Pi(\mu, \nu) = \{ \gamma \in \mathbb{R}^{n \times m}_{+} \mid \gamma \mathbf{1}_m = \mu, \gamma^{\intercal} \mathbf{1}_n = \nu \}$ is the transport polytope.

\paragraph{Entropic Regularization} Solving the standard OT problem in Eq.~\eqref{eq:ot_primal} is computationally prohibitive for large-scale applications. To address this, Cuturi~\cite{cuturi2013sinkhorndistanceslightspeedcomputation} introduced an entropic regularization term $H(\gamma) = - \sum_{i,j} \gamma_{ij} (\log \gamma_{ij} - 1)$, resulting in the strictly convex Sinkhorn distance problem:
\begin{equation}
    \gamma^*_\varepsilon = \argmin_{\gamma \in \Pi(\mu, \nu)} \langle \gamma, C \rangle_F - \varepsilon H(\gamma),
    \label{eq:ot_entropic}
\end{equation}
where $\varepsilon > 0$ is a regularization coefficient. This regularization makes the optimization problem strictly convex, ensuring a unique optimal solution that approximates the original OT problem in Eq.~\eqref{eq:ot_primal}, while enabling efficient computation using the Sinkhorn-Knopp algorithm~\cite{Sinkhorn1967ConcerningNM}.

\paragraph{Unbalanced Optimal Transport} Standard OT enforces strict mass conservation ($\sum \mu_i = \sum \nu_j$), making it sensitive to outliers. In image-text retrieval, where images may contain background clutter or captions may include non-visual words, strict alignment is often suboptimal. Unbalanced Optimal Transport (UOT)~\cite{frogner2015learning} addresses this by relaxing the marginal constraints using divergence penalties, such as the Kullback-Leibler (KL) divergence. The entropic regularized UOT problem is formulated as:
\begin{equation}
    \gamma^*_{\text{UOT}} = \argmin_{\gamma \in \mathbb{R}^{n \times m}_{+}} \langle \gamma, C \rangle_F - \varepsilon H(\gamma) + \tau_{m1} \mathrm{KL}(\gamma \mathbf{1}_m \| \mu) + \tau_{m2} \mathrm{KL}(\gamma^{\intercal} \mathbf{1}_n \| \nu),
    \label{eq:uot}
\end{equation}
where $\tau_{m1}, \tau_{m2} > 0$ control the degree of relaxation for the marginal constraints, and the KL divergence between vectors $\mathbf{u}$ and $\mathbf{v}$ is defined as:
\begin{equation}
    \mathrm{KL}(\mathbf{u} \| \mathbf{v}) = \sum_{i=1}^{n} u_i \log\left(\frac{u_i}{v_i}\right) - u_i + v_i.
    \label{eq:kl_divergence}
\end{equation}
This formulation allows partial matching, enabling the model to handle noisy pairs effectively by ignoring irrelevant visual or textual elements during alignment.

\section{Proposed Method}\label{sec:proposed-method}

\begin{figure}[htbp]
    \centering
    \includegraphics[width=.95\linewidth]{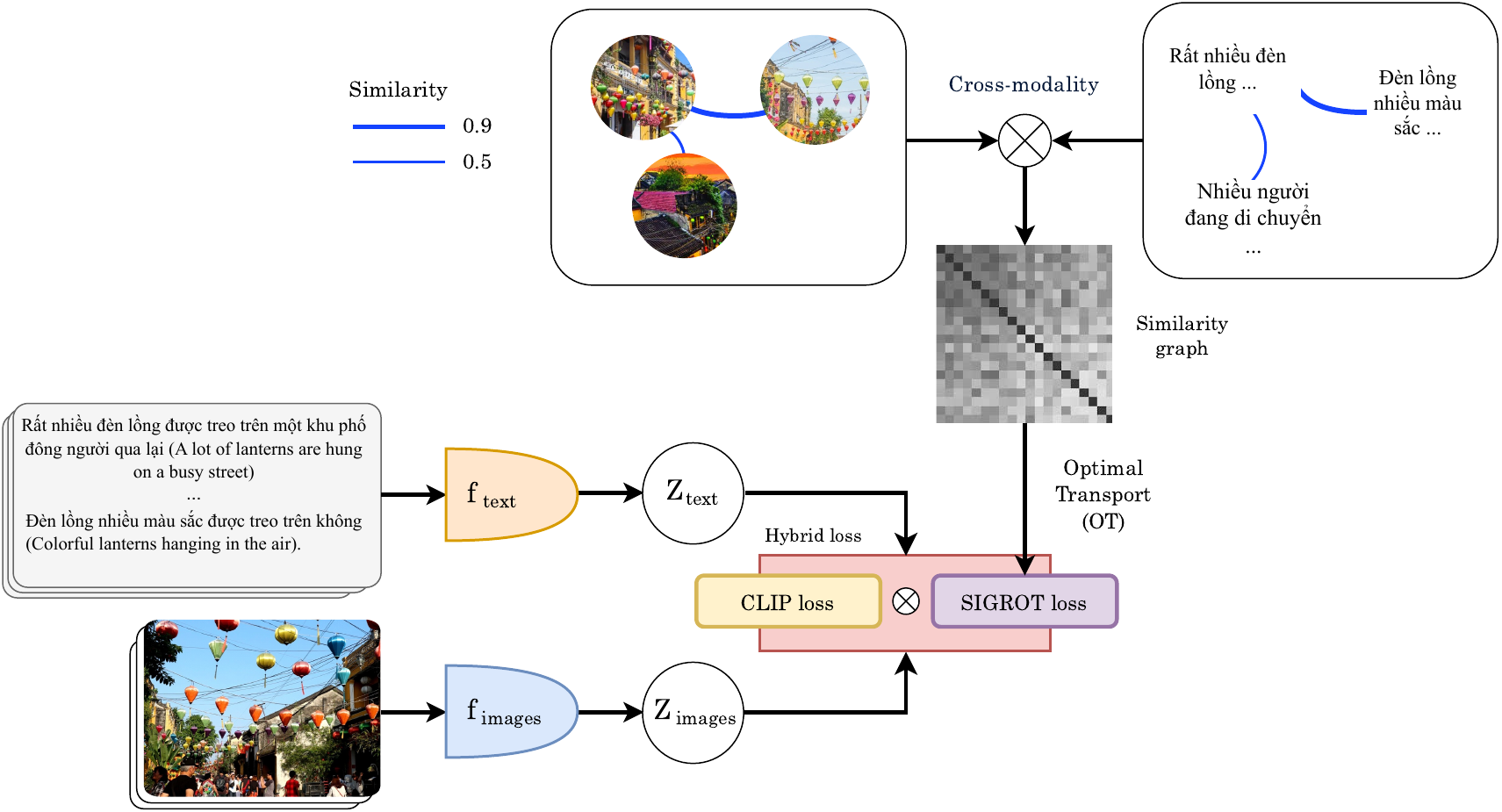}
    \caption{ViCLIP-OT architecture overview. The model consists of a DINOv3-based image encoder and a Vietnamese Sentence-BERT text encoder that project images and texts into a shared embedding space. The hybrid training objective combines a CLIP-style contrastive loss with the proposed SIGROT loss, which uses a similarity graph and optimal transport to enforce global cross-modal alignment.}
    \label{fig:viclip_ot_overview}
\end{figure}

This section describes ViCLIP-OT, a Vietnamese vision--language retrieval model trained with a hybrid objective that combines CLIP-style contrastive learning and an OT-based regularizer. The overall pipeline is depicted in Fig.~\ref{fig:viclip_ot_overview}, where image and text embeddings are aligned by the CLIP loss while the OT module leverages a similarity graph to impose distribution-level cross-modal consistency.

\subsection{Architecture Overview}\label{subsec:architecture}
ViCLIP-OT follows a dual-encoder design with an image tower $f_{\text{image}}$ and a text tower $f_{\text{text}}$. Given an image $x_i$ and a text $t_i$, the encoders output $d$-dimensional embeddings:
\begin{equation}
\mathbf{z}^{\text{image}}_i = f_{\text{image}}(x_i) \in \mathbb{R}^d, \qquad
\mathbf{z}^{\text{text}}_i = f_{\text{text}}(t_i) \in \mathbb{R}^d,
\end{equation}
followed by $\ell_2$ normalization:
\begin{equation}
\tilde{\mathbf{z}} = \frac{\mathbf{z}}{\|\mathbf{z}\|_2}.
\end{equation}

\paragraph{Image tower}
The image encoder is based on DINOv3~\cite{simeoni2025dinov3}, a state-of-the-art ViT pretrained using self-distillation on large-scale image datasets. Given an input image, the DINOv3 backbone extracts patch-level features, which are then aggregated using mean pooling to obtain a global image representation. A linear projection layer maps this representation to the shared $d$-dimensional embedding space.

\paragraph{Text tower}
The text encoder uses a pretrained Vietnamese Sentence-BERT (SBERT) model. Given tokenized inputs, token representations are aggregated via mean pooling over non-padding tokens to obtain a sentence embedding. An optional linear projection is applied when the SBERT hidden size does not match $d$.

The two encoders are jointly trained using a hybrid loss that combines CLIP-style contrastive learning with an OT-based regularizer, as detailed in the following sections.

\subsection{Similarity-Graph Regularized Optimal Transport Loss}\label{subsec:sigrot}
This work proposes SIGROT (\underline{Si}milarity-\underline{G}raph \underline{R}egularized \underline{O}ptimal {\allowbreak}\underline{T}ransport), an OT-based objective designed to complement CLIP-style contrastive learning by injecting \emph{global structure} into batch-wise cross-modal matching. As shown in Fig.~\ref{fig:viclip_ot_overview}, SIGROT constructs a similarity graph and uses OT to regularize the alignment between image and text embeddings.

\paragraph{Motivation}
The standard CLIP loss enforces instance-level alignment using in-batch negatives, but it does not explicitly exploit \emph{relational structure} among samples (e.g., multiple captions describing similar visual concepts). SIGROT addresses this by (i) building a similarity graph that encodes relationships among samples within a batch and (ii) using an OT solver to encourage a globally consistent image--text matching that respects such relationships.

\paragraph{Batch similarity graph construction}\label{para:similarity_graph_construction}
For each training batch, SIGROT uses \emph{precomputed and $\ell_2$-normalized} embeddings to build similarity graphs. These embeddings can be computed using a robust pretrained embedding model. Let $E_{\text{text}}$ and $E_{\text{image}}$ denote the matrices that stack precomputed caption and image embeddings for the samples in the current batch. A text--text similarity matrix and an image--image similarity matrix are computed as:
\begin{equation}
G_{\texttext} = E_{\text{text}} E_{\text{text}}^{\intercal}, \qquad
G_{\imageimage} = E_{\text{image}} E_{\text{image}}^{\intercal}.
\end{equation}

Optionally, cross-modality similarity terms are included:
\begin{equation}
G_{\textimage} = E_{\text{text}} E_{\text{image}}^{\intercal}, \qquad
G_{\imagetext} = E_{\text{image}} E_{\text{text}}^{\intercal}.
\end{equation}

These graphs are combined into a single similarity graph $G$ via an aggregation function $\Phi$:
\begin{equation}
G = \Phi\!\left(G_{\texttext},\,G_{\imageimage},\, G_{\textimage},\, G_{\imagetext}\right).
\end{equation}
In this work, $G$ is obtained by averaging all four similarity matrices:
\begin{equation}
G_{\text{cross}} = \frac{1}{4}\left(G_{\texttext} +G_{\imageimage} + G_{\textimage} + G_{\imagetext}\right).
\label{eq:cross_modal_similarity_graph}
\end{equation}
This formulation is referred to as the \emph{cross-modality similarity graph}, denoted $G_{\text{cross}}$, as it captures both intra-modal and inter-modal relationships within each batch. Alternative combination strategies are explored in Section~\ref{ablation:combine_method_similarity_graph}.

The idea of using a similarity graph as soft targets derived from dataset metadata (e.g., captions in image--caption pairs) has been previously explored by Sobal et al.~\cite{sobal2024mathbbxsamplecontrastivelossimproving}. Building upon this idea, the approach is extended to incorporate both text-to-image and image-to-text relations, as formulated in Eq.~\eqref{eq:cross_modal_similarity_graph}.

\paragraph{OT formulation with similarity-graph regularization}
Given model outputs for a batch of size $N$, normalized embeddings $\tilde{Z}_{\text{image}}$ and $\tilde{Z}_{\text{text}}$ are obtained, and their cross-modal similarity matrix is:
\begin{equation}
S_{\imagetext} = \tilde{Z}_{\text{image}} \tilde{Z}_{\text{text}}^{\intercal}.
\end{equation}
The cost matrix for transport is defined as $C_{\imagetext} = \mathbf{1} - S_{\imagetext}$, where higher similarity corresponds to lower transport cost. SIGROT defines a transport plan $\gamma \in \mathbb{R}_{+}^{N\times N}$ that matches images to texts at the batch level. For the image-to-text direction, the optimal transport plan is obtained by solving:
\begin{equation}
\begin{aligned}
    \gamma^{*}_{\text{i2t}} = \argmin_{\gamma \in \mathbb{R}^{N \times N}_{+}} \; & \langle \gamma, C_{\imagetext} \rangle_F - \varepsilon H(\gamma) \\
    & + \tau_{m1} \mathrm{KL}(\gamma \mathbf{1}_N \| \mu) + \tau_{m2} \mathrm{KL}(\gamma^{\intercal} \mathbf{1}_N \| \nu),
\end{aligned}
\label{eq:sigrot_plan_i2t}
\end{equation}
where $\mu = \nu = \frac{1}{N}\mathbf{1}_N$ are uniform distributions over the batch, $\varepsilon > 0$ is the entropic regularization coefficient, and $\tau_{m1}, \tau_{m2} > 0$ control the relaxation of marginal constraints. This formulation corresponds to the standard Unbalanced OT problem (Eq.~\eqref{eq:uot}). Given the optimal transport plan $\gamma^{*}_{\text{i2t}}$, the image-to-text SIGROT loss measures the divergence between the normalized transport plan and the similarity-graph distribution:

\begin{equation}
    \mathcal{L}_{\text{SIGROT}}^{\text{i2t}} = \mathrm{KL}(N\gamma^{*}_{\text{i2t}} \| \mathrm{softmax}(G_{\text{cross}})).
\label{eq:sigrot_i2t}
\end{equation}
Here, the transport plan is scaled by the batch size $N$ so that $N\gamma^{*}_{\text{i2t}}$ sums to one, forming a valid probability distribution. Similarly, $\mathrm{softmax}(G_{\text{cross}})$ normalizes the similarity graph into a probability distribution, enabling a principled KL divergence comparison between the two.

Similarly, for the text-to-image direction, the optimal transport plan is obtained by:
\begin{equation}
\begin{aligned}
    \gamma^{*}_{\text{t2i}} = \argmin_{\gamma \in \mathbb{R}^{N \times N}_{+}} \; & \langle \gamma, C_{\imagetext}^{\intercal} \rangle_F - \varepsilon H(\gamma) \\
    & + \tau_{m1} \mathrm{KL}(\gamma \mathbf{1}_N \| \nu) + \tau_{m2} \mathrm{KL}(\gamma^{\intercal} \mathbf{1}_N \| \mu),
\end{aligned}
\label{eq:sigrot_plan_t2i}
\end{equation}
and the text-to-image SIGROT loss is:
\begin{equation}
    \mathcal{L}_{\text{SIGROT}}^{\text{t2i}} = \mathrm{KL}(N\gamma^{*}_{\text{t2i}} \| \mathrm{softmax}(G_{\text{cross}})).
\label{eq:sigrot_t2i}
\end{equation}

The final SIGROT loss is the average of both directions:
\begin{equation}
    \mathcal{L}_{\text{SIGROT}} = \frac{1}{2}\left(\mathcal{L}_{\text{SIGROT}}^{\text{i2t}} + \mathcal{L}_{\text{SIGROT}}^{\text{t2i}}\right).
\label{eq:sigrot_loss}
\end{equation}

\subsection{Hybrid Training Objective}\label{subsec:hybrid-training-objective}
ViCLIP-OT is optimized with a hybrid objective (Fig.~\ref{fig:viclip_ot_overview}) that combines a contrastive loss with the proposed SIGROT loss. Two contrastive learning formulations are adopted, namely the standard CLIP loss~\cite{pmlr-v139-radford21a} and the SigLIP loss~\cite{zhai2023sigmoid}, resulting in two hybrid objective functions:
\begin{equation}
\mathcal{L}_{\text{CLIP-SIGROT}} = \lambda \,\mathcal{L}_{\text{CLIP}} + \mathcal{L}_{\text{SIGROT}},
\label{eq:clip_sigrot_loss}
\end{equation}
\begin{equation}
\mathcal{L}_{\text{SigLIP-SIGROT}} = \lambda \,\mathcal{L}_{\text{SigLIP}} + \mathcal{L}_{\text{SIGROT}},
\label{eq:siglip_sigrot_loss}
\end{equation}
where $\lambda \geq 0$ is a hyperparameter that balances the two objectives.

The CLIP loss is a symmetric cross-entropy objective that pulls matched image--text pairs together while pushing mismatched pairs apart within each batch. It applies softmax normalization over all samples, making the learning signal dependent on the entire batch. The SigLIP loss replaces the softmax-based formulation with a sigmoid loss that operates on individual image--text pairs independently. This formulation removes the need for global normalization and enables more efficient training with larger batch sizes. Despite their differences, both CLIP and SigLIP treat each sample independently and do not explicitly model relationships among samples within a batch. SIGROT addresses this limitation by using a precomputed similarity graph and optimal transport to find globally consistent matchings across the batch. Combining a contrastive loss (CLIP or SigLIP) with SIGROT allows the model to learn strong pairwise alignment while also capturing the relational structure among samples.

\paragraph{Stabilizing the training for CLIP and SigLIP}
Following~\cite{pmlr-v139-radford21a,wu2018unsupervised,zhai2023sigmoid}, two learnable parameters are introduced to control the range of the logits: a temperature $\tau$ and a bias $b$. The temperature is parameterized as $\tau = \exp(\tau')$, where $\tau'$ is a learnable scalar, ensuring positivity throughout training. The bias term $b$ is applied only in the SigLIP formulation. Together, these parameters help stabilize optimization by adaptively adjusting the logit distribution.

In all experiments, ViCLIP-OT and ViSigLIP-OT are trained using the hybrid objectives defined in Eqs.~\eqref{eq:clip_sigrot_loss} and~\eqref{eq:siglip_sigrot_loss}, respectively. For brevity, subsequent sections describe the methodology in terms of ViCLIP-OT; the corresponding results for ViSigLIP-OT are included alongside in Section~\ref{sec:results}.

\section{Experiments Settings}\label{sec:experiments}
\subsection{Datasets}
\label{subsec:datasets}
ViCLIP-OT is trained and evaluated using three datasets: UIT-OpenViIC, KTVIC, and Crossmodal-3600. UIT-OpenViIC serves as the primary dataset for training and in-domain evaluation, while KTVIC and Crossmodal-3600 (or \crossmodaldataset) are used exclusively for zero-shot evaluation to assess cross-dataset generalization. To prevent train--test contamination, near-duplicate images in the evaluation datasets are identified and removed against the UIT-OpenViIC training set using SSCD~\cite{pizzi2022self} (see~\ref{app:remove-near-duplicate} for details). Detailed descriptions of each dataset are provided in~\ref{app:dataset-details}.

\paragraph{UIT-OpenViIC}
UIT-OpenViIC~\cite{BUI2026117430} is a large-scale open-domain Vietnamese image captioning dataset comprising 13,100 images with 61,241 captions, split into training (9,088 images), validation (2,011 images), and test (2,001 images) sets.

\paragraph{KTVIC}
KTVIC~\cite{pham2024ktvicvietnameseimagecaptioning} is a Vietnamese image captioning benchmark containing 4,327 images with 21,635 captions depicting daily-life activities and locations in Vietnam. Near-duplicate detection against the UIT-OpenViIC training set reveals substantial overlap: 401 out of 558 test images and 2,464 out of 3,769 training images are identified as near-duplicates. After removal, the deduplicated test split contains 157 images and the training split contains 1,305 images. Due to the severely reduced test set, zero-shot evaluation is performed on both splits. For the remainder of this paper, KTVIC-train and KTVIC-test refer to the deduplicated splits unless otherwise stated.

\paragraph{Crossmodal-3600}
Crossmodal-3600 (\crossmodaldataset)~\cite{thapliyal2022crossmodal} is a geographically diverse multilingual dataset of 3,600 images with captions in 36 languages. No near-duplicates against the UIT-OpenViIC training set are found. All 3,600 images paired with Vietnamese captions are used for zero-shot evaluation, yielding 7,350 image--caption pairs.

\subsection{Evaluation metrics}
\label{subsec:evaluation-metrics}
The performance of ViCLIP-OT on the image--text retrieval task is evaluated using \text{Recall@K} (R@K), a standard metric for cross-modal retrieval~\cite{karpathy2015deep,chen2020uniter,pmlr-v139-radford21a,zhai2022litzeroshottransferlockedimage,zhai2023sigmoid,zhan2025elip}. R@K measures the proportion of queries for which at least one relevant item is retrieved within the top-$K$ ranked results.

\paragraph{Text Retrieval}
In the text retrieval setting, each image is associated with multiple ground-truth captions. Given an image query, the model ranks all candidate captions based on their similarity scores. The retrieval is considered successful if \emph{any} of the corresponding ground-truth captions appear within the top-$K$ retrieved captions. R@K is then computed as the percentage of image queries that satisfy this condition.

\paragraph{Image Retrieval}
In the image retrieval setting, each caption is paired with exactly one ground-truth image. Given a caption query, the model ranks all candidate images. A retrieval is considered correct if the associated ground-truth image is retrieved within the top-$K$ ranked images. R@K is calculated as the percentage of caption queries for which this criterion holds.

\paragraph{Embedding Space Quality}
To assess the quality of the learned shared embedding space, two additional metrics are used. The \textit{Alignment score}~\cite{goel2022cyclip} measures how well matched image-text pairs are aligned:
\begin{equation}
    \text{Alignment} = \frac{1}{N} \sum_{i=1}^{N} \text{sim}(\mathbf{z}^{\text{image}}_i, \mathbf{z}^{\text{text}}_i),
\end{equation}
where $\mathrm{sim}(\cdot, \cdot)$ denotes the cosine similarity and higher alignment scores indicate better pairwise alignment. The \textit{Modality gap}~\cite{liang2022mind} quantifies the distance between the centroids of image and text embeddings:
\begin{equation}
    \Delta_{\text{gap}} = \frac{1}{N} \sum_{i=1}^{N} \mathbf{z}^{\text{image}}_i - \frac{1}{N} \sum_{i=1}^{N} \mathbf{z}^{\text{text}}_i,
\end{equation}
where lower values indicate smaller separation between modalities.

\subsection{Baseline Models}
The baseline models follow the standard dual-encoder architecture for image--text retrieval~\cite{pmlr-v139-radford21a,zhai2023sigmoid, zhai2021lit}. For the visual modality, a Vision Transformer (ViT-B/16) backbone pretrained using the DINOv3~\cite{simeoni2025dinov3} self-supervised learning framework\footnote{\url{https://huggingface.co/timm/vit_base_patch16_dinov3.lvd1689m}} is employed to encode images into fixed-dimensional embeddings. For the textual modality, an SBERT~\cite{reimers-2019-sentence-bert} model that has been previously fine-tuned on Vietnamese corpora\footnote{\url{https://huggingface.co/keepitreal/vietnamese-sbert}} is used to encode text descriptions. Both image and text encoders project their respective inputs into a shared embedding space, where cross-modal similarity is computed using cosine similarity.

Two baseline models are constructed with an identical encoder architecture, while differing in their respective training objectives:
\begin{itemize}
    \item CLIP baseline: The model is trained using the original CLIP contrastive loss~\cite{pmlr-v139-radford21a}, which applies softmax normalization over all samples in a batch and encourages matched image--text pairs to have higher similarity scores than mismatched pairs.
    \item SigLIP baseline: The model is trained using the SigLIP loss~\cite{zhai2023sigmoid}, which replaces softmax-based normalization with a sigmoid loss that operates on individual image--text pairs independently, enabling more efficient training with larger batch sizes.
\end{itemize}

All parameters of both backbone encoders are jointly optimized during training, enabling cross-modal alignment to be learned directly from Vietnamese image--text data.

\subsection{Implementation Details}
ViCLIP-OT is implemented using PyTorch\footnote{\url{https://pytorch.org}}
 with Python 3.12. All experiments are conducted on a single NVIDIA RTX 4090 GPU with 24 GB of memory. To facilitate large-batch contrastive training under limited GPU memory, gradient accumulation is combined with the gradient caching strategy proposed by Gao et al.~\cite{gao2021scaling}, enabling an effectively larger batch size without incurring additional GPU memory overhead.

\paragraph{Pre-computing embeddings for the similarity graph}
To construct the similarity graph described in Section~\ref{subsec:sigrot}, Qwen3-VL-Embedding-2B~\cite{qwen3vlembedding} is employed as a compact yet robust multimodal multilingual embedding model capable of encoding both images and text. Embeddings are pre-computed for all image--text pairs in the training set and subsequently used to compute pairwise cosine similarities, which form the graph $G_{\text{cross}}$ in Eq.~\eqref{eq:cross_modal_similarity_graph}.

\paragraph{Training configuration}
Images are resized to $224 \times 224$ pixels and normalized using ImageNet statistics. Data augmentation includes random rotation, random horizontal flip, and color jitter (brightness$=$0.2, contrast$=$0.2, saturation$=$0.2, hue$=$0.0). The embedding dimension is set to $768$ for both image and text encoders. All models are trained for 30 epochs with a batch size of 128 using mixed-precision training with \texttt{bfloat16}. Optimization is performed using the AdamW optimizer~\cite{loshchilov2017decoupled}, configured with $\epsilon = 10^{-10}$, $\beta = (0.9, 0.999)$, and a weight decay coefficient of $10^{-4}$. A differentiated learning rate strategy is adopted: the backbone encoders use a peak learning rate of $5 \times 10^{-5}$, while the linear projection heads and MLPs use $2 \times 10^{-4}$. The learning rate follows a warmup-cosine schedule: linearly warming up from $0.01 \times \texttt{peak\_lr}$ over the first 2 epochs, then decaying to $0.0001 \times \texttt{peak\_lr}$ using cosine annealing. Gradient clipping is applied with a maximum norm of 1.0. For the hybrid training objectives in Eq.~\eqref{eq:clip_sigrot_loss} and Eq.~\eqref{eq:siglip_sigrot_loss}, the balancing coefficient is set to $\lambda = 0.1$. Following~\cite{pmlr-v139-radford21a, zhai2023sigmoid}, the temperature and bias parameters are initialized as $\tau = 0.07$ for CLIP, and $\tau = 0.06$, $b = -9$ for SigLIP. The regularization coefficients in Eq.~\eqref{eq:sigrot_plan_i2t} and Eq.~\eqref{eq:sigrot_plan_t2i} are set to $\varepsilon = 0.05$, $\tau_{m1} = \tau_{m2} = 0.5$. Since each image may be associated with multiple captions, all corresponding captions are included in the batch alongside their paired image.

\paragraph{Model selection}
The final model for evaluation on the test set is selected based on the checkpoint that achieves the highest text-to-image R@1 on the validation set.

\section{Results}\label{sec:results}
This section presents the experimental results of ViCLIP-OT on image--text retrieval tasks, including in-domain evaluation on UIT-OpenViIC and zero-shot evaluation on KTVIC and Crossmodal-3600 datasets. The performance of ViCLIP-OT is compared against the baseline CLIP and SigLIP models and other relevant methods to demonstrate the effectiveness of the proposed approach.

\subsection{Image-Text Retrieval on UIT-OpenViIC}

\begin{figure*}[tbp]
    \centering
    \begin{minipage}[t]{0.48\textwidth}
        \centering
        \begin{tikzpicture}
        \begin{axis}[
            ybar,
            bar width=8pt,
            width=\textwidth,
            height=6cm,
            xlabel={\small Recall@K},
            ylabel={\small Value (\%)},
            title={Text-to-Image},
            symbolic x coords={R@1, R@5, R@10},
            xtick=data,
            ymin=25, ymax=82,
            ytick={30, 40, 50, 60, 70, 80},
            grid=major,
            grid style=dashed,
            enlarge x limits=0.2,
        ]
        \addplot[
            draw=blue!100!black,
            pattern=north east lines,
            pattern color=blue!100!black,
        ] coordinates {
            (R@1, 31.19) (R@5, 59.80) (R@10, 71.23)
        };
        \addplot[
            draw=orange!100!black,
            pattern=crosshatch dots,
            pattern color=orange!100!black,
        ] coordinates {
            (R@1, 34.75) (R@5, 63.01) (R@10, 72.96)
        };
        \addplot[
            draw=red!100!black,
            pattern=horizontal lines,
            pattern color=red!100!black,
        ] coordinates {
            (R@1, 37.57) (R@5, 65.65) (R@10, 75.43)
        };
        \addplot[
            draw=green!75!black,
            pattern=north west lines,
            pattern color=green!75!black,
        ] coordinates {
            (R@1, 39.19) (R@5, 66.71) (R@10, 76.04)
        };
        \end{axis}
        \end{tikzpicture}
    \end{minipage}
    \hfill
    \begin{minipage}[t]{0.48\textwidth}
        \centering
        \begin{tikzpicture}
        \begin{axis}[
            ybar,
            bar width=8pt,
            width=\textwidth,
            height=6cm,
            xlabel={\small Recall@K},
            ylabel={\small Value (\%)},
            title={Image-to-Text},
            symbolic x coords={R@1, R@5, R@10},
            xtick=data,
            ymin=40, ymax=95,
            ytick={45, 55, 65, 75, 85, 95},
            grid=major,
            grid style=dashed,
            enlarge x limits=0.2,
            legend to name=legendOpenViIC,
            legend columns=4,
            legend style={font=\footnotesize, draw=none},
        ]
        \addplot[
            draw=blue!100!black,
            pattern=north east lines,
            pattern color=blue!100!black,
        ] coordinates {
            (R@1, 46.60) (R@5, 75.53) (R@10, 85.19)
        };
        \addplot[
            draw=orange!100!black,
            pattern=crosshatch dots,
            pattern color=orange!100!black,
        ] coordinates {
            (R@1, 50.10) (R@5, 79.78) (R@10, 88.04)
        };
        \addplot[
            draw=red!100!black,
            pattern=horizontal lines,
            pattern color=red!100!black,
        ] coordinates {
            (R@1, 54.35) (R@5, 81.83) (R@10, 89.19)
        };
        \addplot[
            draw=green!75!black,
            pattern=north west lines,
            pattern color=green!75!black,
        ] coordinates {
            (R@1, 57.21) (R@5, 83.83) (R@10, 90.79)
        };
        \legend{CLIP, SigLIP, ViCLIP-OT, ViSigLIP-OT}
        \end{axis}
        \end{tikzpicture}
    \end{minipage}
    \\[0.5em]
    \pgfplotslegendfromname{legendOpenViIC}
    \caption{R@K comparison on UIT-OpenViIC for text-to-image (left) and image-to-text (right) retrieval tasks. Incorporating the SIGROT loss consistently improves performance over both CLIP and SigLIP baselines across all R@K metrics.}
    \label{fig:uit-openviic-retrieval}
\end{figure*}
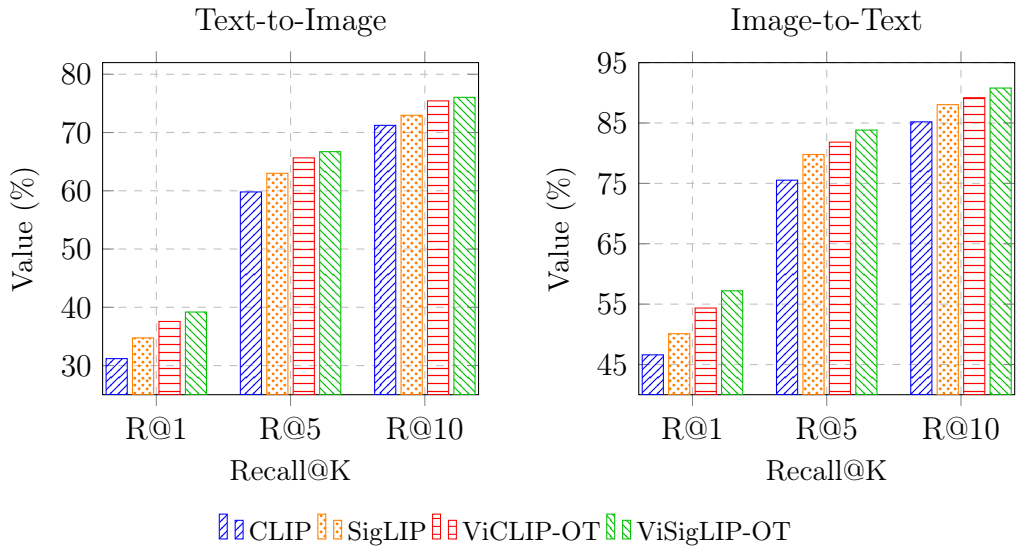

The retrieval performance of ViCLIP-OT on the UIT-OpenViIC test set is summarized in Table~\ref{tab:uit-openvic-retrieval}. The results demonstrate that ViCLIP-OT significantly outperforms the baseline CLIP and SigLIP models across all R@K metrics for both text-to-image and image-to-text retrieval tasks. Specifically, ViCLIP-OT achieves an average R@K of 67.34\%, compared to 61.59\% for the baseline CLIP model, representing a substantial improvement of 5.75 percentage points. While ViSigLIP-OT attains an average R@K of 68.96\%, surpassing ViCLIP-OT by 1.62 percentage points. This performance gain highlights the effectiveness of incorporating the SIGROT loss, which leverages optimal transport and similarity graph to enhance cross-modal alignment. A visual comparison of R@K across models is presented in Figure~\ref{fig:uit-openviic-retrieval}.



\begin{table*}[tbp]
\centering
\fontsize{12pt}{12pt}\selectfont
\caption{Image-text retrieval performance on the test set of the UIT-OpenViIC dataset. UOT denotes Unbalanced Optimal Transport. * indicates zero-shot evaluation. Best results are in \textbf{bold} and second-best are \underline{underlined}. ViCLIP-OT and ViSigLIP-OT outperform their corresponding baselines (CLIP and SigLIP without SIGROT), other loss variants (CLIP + UOT, SigLIP + UOT, SIGROT), and pretrained vision--language models under zero-shot settings, demonstrating the effectiveness of integrating the SIGROT loss.}
\label{tab:uit-openvic-retrieval}
\resizebox{\textwidth}{!}{%
\begin{tabular}{l c ccc ccc | c}
\toprule
\multirow{2}{*}{Method/Model} & \multirow{2}{*}{\texttt{\#} Params} & \multicolumn{3}{c}{Text $\rightarrow$ Image} & \multicolumn{3}{c|}{Image $\rightarrow$ Text} & \multirow{2}{*}{\makebox[1.05cm]{Avg.}} \\
\cmidrule(lr){3-5} \cmidrule(lr){6-8}
& & \makebox[1.05cm]{R@1} & \makebox[1.05cm]{R@5} & \makebox[1.05cm]{R@10} & \makebox[1.05cm]{R@1} & \makebox[1.05cm]{R@5} & \makebox[1.05cm]{R@10} & \\
\midrule
mSigLIP-base*~\cite{zhai2023sigmoid} & 370M & 14.34 & 28.94 & 36.21 & 20.49 & 32.23 & 37.43 & 28.27 \\
Jina CLIP v2*~\cite{koukounas2024jinaclipv2multilingualmultimodalembeddings} & 865M & 30.01 & 52.09 & 61.70 & 40.23 & 65.02 & 74.41 & 53.91 \\
Jina Embedding v4*~\cite{gunther2025jina} & 4B & 23.97 & 42.22 & 50.29 & 41.48 & 66.77 & 75.61 & 50.06 \\
Qwen3-VL-Embedding-2B*~\cite{qwen3vlembedding} & 2B & 32.13 & 54.00 & 62.93 & 39.83 & 66.52 & 77.01 & 55.40 \\
\midrule
CLIP & 221M & 31.19 & 59.80 & 71.23 & 46.60 & 75.53 & 85.19 & 61.59 \\
SigLIP & 221M & 34.75 & 63.01 & 72.96 & 50.10 & 79.78 & 88.04 & 64.77 \\
\midrule
CLIP + UOT & 221M & 29.27 & 57.62 & 69.07 & 43.59 & 75.03 & 84.03 & 59.77 \\
SigLIP + UOT & 221M & 37.84 & 65.30 & 74.98 & 53.95 & 80.95 & 88.81 & 66.97 \\
SIGROT & 221M & \textbf{40.75} & \textbf{70.72} & \textbf{80.90} & 37.99 & 61.11 & 71.68 & 60.53 \\
ViCLIP-OT (Eq.~\ref{eq:clip_sigrot_loss}) & 221M & 37.57 & 65.65 & 75.43 & \underline{54.35} & \underline{81.83} & \underline{89.19} & \underline{67.34} \\
ViSigLIP-OT (Eq.~\ref{eq:siglip_sigrot_loss}) & 221M & \underline{39.19} & \underline{66.71} & \underline{76.04} & \textbf{57.21} & \textbf{83.83} & \textbf{90.79} & \textbf{68.96} \\
\bottomrule
\end{tabular}
}%
\end{table*}

Beyond the baseline CLIP-style model, ViCLIP-OT is compared against several pretrained multilingual vision--language models in a zero-shot setting, including mSigLIP~\cite{zhai2023sigmoid}, Jina CLIP v2~\cite{koukounas2024jinaclipv2multilingualmultimodalembeddings}, Jina Embedding v4~\cite{gunther2025jina}, and Qwen3-VL-Embedding-2B~\cite{qwen3vlembedding}. ViCLIP-OT outperforms all these models by a significant margin, achieving an average R@K of 67.34\%, compared to 55.40\% for Qwen3-VL-Embedding-2B, representing an improvement of 11.94 percentage points. ViSigLIP-OT further enhances performance, achieving an average R@K of 68.96\%, surpassing Qwen3-VL-Embedding-2B by 13.56 percentage points. This result underscores the advantage of training on domain-specific Vietnamese image--text data and the effectiveness of the proposed SIGROT loss in enhancing cross-modal retrieval performance. Additional experiments in \ref{app:retrieval-results-with-different-model-architectures} further confirm that SIGROT consistently improves performance across different model architectures. A qualitative comparison between SigLIP and ViSigLIP-OT with retrieval examples is provided in \ref{app:retrieval-examples}.


\subsection{Zero-shot Image-Text Retrieval}

The zero-shot retrieval performance of ViCLIP-OT on the KTVIC and Crossmodal-3600 datasets is presented in Table~\ref{tab:zero_shot_retrieval}. As described in Section~\ref{subsec:datasets}, near-duplicate images in KTVIC are removed against the UIT-OpenViIC training set to ensure fair evaluation (see \ref{app:remove-near-duplicate}). Due to the substantial overlap (401 out of 558 test images), evaluation is conducted on both the deduplicated training and test splits of KTVIC. The results indicate that ViCLIP-OT significantly outperforms both the baseline CLIP model and the SigLIP variant across all R@K metrics in both text-to-image and image-to-text retrieval tasks. On the KTVIC test split, ViCLIP-OT achieves an average R@K of 82.68\%, compared to 79.32\% for CLIP, representing an improvement of 3.36 percentage points. On Crossmodal-3600, ViCLIP-OT attains an average R@K of 56.85\%, surpassing CLIP's 45.13\% by a substantial margin of 11.72 percentage points. Alongside ViCLIP-OT, ViSigLIP-OT also demonstrates strong performance, achieving an average R@K of 54.37\% on KTVIC train split, 82.78\% on KTVIC test split, and 56.17\% on Crossmodal-3600. These findings underscore the effectiveness of the SIGROT loss in enhancing cross-modal retrieval capabilities, particularly in zero-shot scenarios where models must generalize to unseen data distributions.

\begin{table*}[tbp]
\centering
\small
\caption{Zero-shot image--text retrieval results on KTVIC and Crossmodal-3600. KTVIC images are deduplicated against the UIT-OpenViIC training set (see \ref{app:remove-near-duplicate}). Vietnamese captions are used for Crossmodal-3600. ViCLIP-OT and ViSigLIP-OT consistently outperform their corresponding baselines across both datasets in zero-shot settings, confirming the generalizability of the SIGROT loss.}
\label{tab:zero_shot_retrieval}
\resizebox{0.9\textwidth}{!}{%
\begin{tabular}{l ccc ccc | c}
\toprule
\multirow{2}{*}{Method} & \multicolumn{3}{c}{Text $\rightarrow$ Image} & \multicolumn{3}{c|}{Image $\rightarrow$ Text} & \multirow{2}{*}{\makebox[1.05cm]{Avg.}} \\
\cmidrule(lr){2-4} \cmidrule(lr){5-7}
& \makebox[1.05cm]{R@1} & \makebox[1.05cm]{R@5} & \makebox[1.05cm]{R@10} & \makebox[1.05cm]{R@1} & \makebox[1.05cm]{R@5} & \makebox[1.05cm]{R@10} & \\
\midrule
\multicolumn{8}{c}{\textit{KTVIC-train}} \\
\midrule
CLIP & 21.12 & 46.99 & 59.22 & 31.65 & 59.46 & 72.49 & 48.49 \\
SigLIP & 23.16 & 48.78 & 60.57 & 35.48 & 62.22 & 73.64 & 50.64 \\
ViCLIP-OT & 26.24 & 52.46 & \textbf{64.14} & 38.47 & 64.37 & 75.48 & 53.52 \\
ViSigLIP-OT & \textbf{26.28} & \textbf{52.58} & 63.49 & \textbf{39.62} & \textbf{66.44} & \textbf{77.78} & \textbf{54.37} \\
\midrule
\multicolumn{8}{c}{\textit{KTVIC-test}} \\
\midrule
CLIP & 50.32 & 82.80 & 89.94 & 63.06 & 92.36 & 97.45 & 79.32 \\
SigLIP & 52.61 & 83.31 & 89.94 & \textbf{71.97} & \textbf{94.27} & 96.18 & 81.38 \\
ViCLIP-OT & \textbf{56.69} & 85.61 & \textbf{91.97} & 70.06 & 93.63 & \textbf{98.09} & 82.68 \\
ViSigLIP-OT & 56.56 & \textbf{85.99} & 91.72 & 71.34 & 93.63 & 97.45 & \textbf{82.78} \\
\midrule
\multicolumn{8}{c}{\textit{Crossmodal-3600}} \\
\midrule
CLIP & 22.52 & 45.55 & 58.01 & 26.22 & 53.42 & 65.06 & 45.13 \\
SigLIP & 26.67 & 50.31 & 61.78 & 31.17 & 57.78 & 69.83 & 49.59 \\
ViCLIP-OT & 28.90 & 55.29 & 66.37 & \textbf{42.56} & \textbf{68.81} & \textbf{79.17} & \textbf{56.85} \\
ViSigLIP-OT & \textbf{32.04} & \textbf{57.90} & \textbf{68.95} & 37.97 & 64.64 & 75.53 & 56.17 \\
\bottomrule
\end{tabular}
}%
\end{table*}

\subsection{Visualization of Embedding Space}\label{subsec:visualization-of-embedding-space}
\begin{figure}[htbp]
    \centering
    \includegraphics[width=1.025\linewidth]{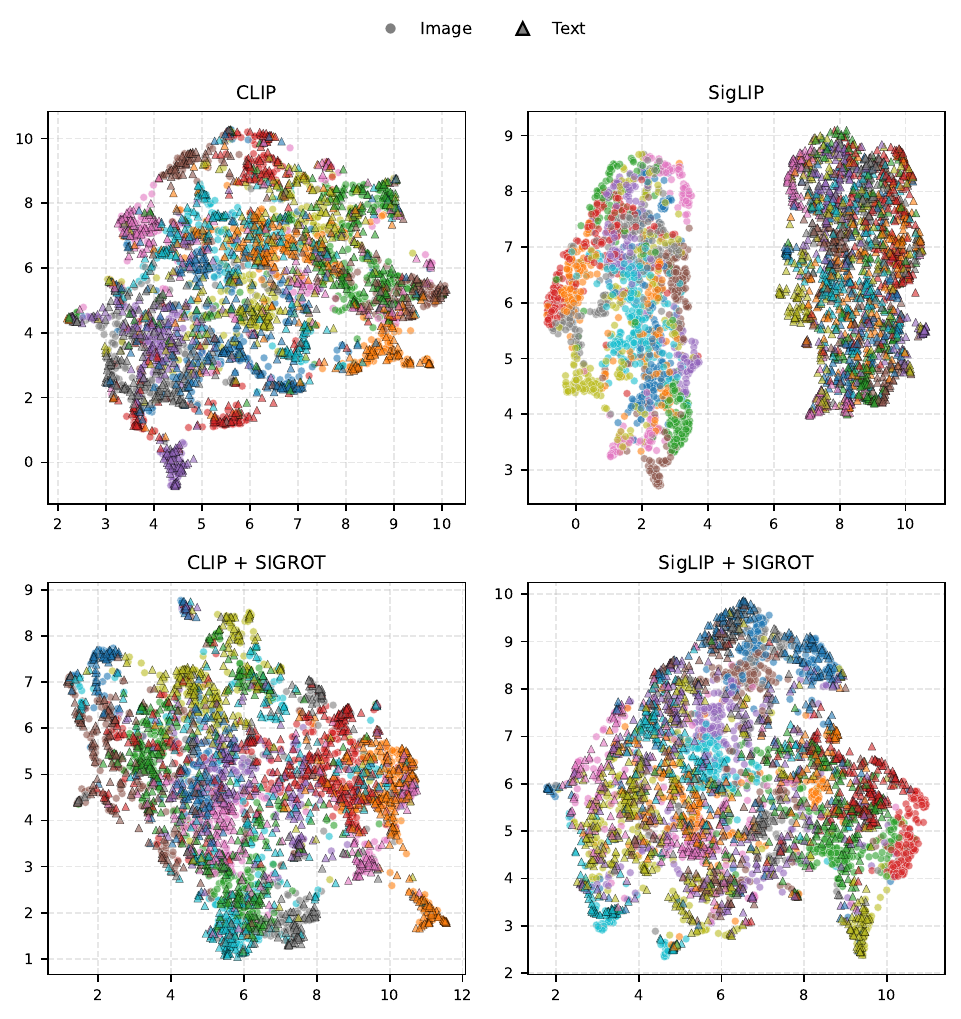}
    \caption{UMAP visualization of image and text embeddings on the UIT-OpenViIC test set. Each subplot corresponds to a different training objective. Circles represent image embeddings and triangles represent text embeddings, with colors indicating pseudo labels obtained via K-Means clustering ($k=20$). SIGROT-based methods exhibit tighter cross-modal clustering compared to baselines.}
    \label{fig:viz-embeddings-space}
\end{figure}

To better understand the learned embedding spaces, UMAP~\cite{mcinnes2018umap} is used to visualize image and text embeddings from models trained with different objectives. Figure~\ref{fig:viz-embeddings-space} shows these visualizations, where circles denote image embeddings and triangles denote text embeddings.

The \textit{modality gap}~\cite{liang2022mind} is a known issue in multimodal representation learning, where different modalities occupy distinct regions within the shared latent space (for more details, refer to \ref{app:modality-gap}). This separation is commonly observed in models trained with cross-modal contrastive objectives, such as those used in CLIP and SigLIP. The modality gap can hinder cross-modal retrieval performance, as semantically similar image-text pairs may not be sufficiently close in the embedding space.

In Figure~\ref{fig:viz-embeddings-space}, the modality gap is clearly visible in SigLIP, where image and text embeddings form two separate clusters with little overlap. CLIP, by contrast, exhibits no apparent modality gap, with the two modalities well interleaved. This difference is hypothesized to stem from the sensitivity of SigLIP to its temperature and bias parameters during the early stages of training, though a thorough investigation of how these parameters influence the modality gap is outside the scope of this work. Notably, adding SIGROT improves cross-modal alignment in both cases. For ViSigLIP-OT, some misalignment remains compared to ViCLIP-OT, but the gap is substantially reduced relative to SigLIP alone. ViCLIP-OT shows the tightest clustering, with matched pairs closely co-located, indicating that SIGROT enforces geometric consistency between modalities.

To quantify this effect, two metrics defined in Section~\ref{subsec:evaluation-metrics} are used: the \textit{Alignment score} and the \textit{Modality gap}. Results are shown in Table~\ref{tab:alignment-score-modality-gap}.

Table~\ref{tab:alignment-score-modality-gap} shows that SIGROT improves both metrics across all datasets. On UIT-OpenViIC, SigLIP's Alignment score increases from 0.3637 to 0.3928 with SIGROT, while the Modality gap drops from 0.5843 to 0.3177. Similar patterns hold for CLIP-based models: ViCLIP-OT achieves the lowest Modality gap (0.1026 on UIT-OpenViIC), while SIGROT alone achieves the highest Alignment scores. These results show that SIGROT produces tighter coupling between image and text modalities, contributing to improved retrieval performance.

\begin{table*}[htbp]
    \centering
    \caption{Alignment score (A) and Modality gap ($\| \Delta_{\text{gap}} \|$) across different datasets. Higher Alignment scores and lower Modality gaps indicate better cross-modal alignment. SIGROT achieves the highest Alignment scores across all datasets. Integrating SIGROT consistently improves both metrics compared to the corresponding baselines, confirming its role in enhancing cross-modal alignment.}
    \label{tab:alignment-score-modality-gap}
    \resizebox{0.9\textwidth}{!}{%
    \begin{tabular}{l cc cc cc}
    \toprule
    \multirow{2}{*}{Method} & \multicolumn{2}{c}{UIT-OpenViIC} & \multicolumn{2}{c}{KTVIC-test} & \multicolumn{2}{c}{Crossmodal-3600} \\
    \cmidrule(lr){2-3} \cmidrule(lr){4-5} \cmidrule(lr){6-7}
    & \makebox[1.05cm]{A $\uparrow$} & \makebox[1.05cm]{$\| \Delta_{\text{gap}} \| \downarrow$} & \makebox[1.05cm]{A $\uparrow$} & \makebox[1.05cm]{$\| \Delta_{\text{gap}} \| \downarrow$} & \makebox[1.05cm]{A $\uparrow$} & \makebox[1.05cm]{$\| \Delta_{\text{gap}} \| \downarrow$} \\
    \midrule
    SIGROT & \textbf{0.8061} & \underline{0.1323} & \textbf{0.7670} & 0.2135 & \textbf{0.6976} & \underline{0.1625} \\
    \midrule
    CLIP & 0.5201 & 0.1952 & 0.4696 & \underline{0.2032} & 0.5329 & 0.2558 \\
    ViCLIP-OT & \underline{0.6624} & \textbf{0.1026} & \underline{0.6212} & \textbf{0.1636} & \underline{0.6225} & \textbf{0.1273} \\
    \midrule
    SigLIP & 0.3637 & 0.5843 & 0.3182 & 0.5757 & 0.3790 & 0.5789 \\
    ViSigLIP-OT & 0.3928 & 0.3177 & 0.3373 & 0.3385 & 0.4142 & 0.3442 \\
    \bottomrule
    \end{tabular}
    }%
\end{table*}

\subsection{Visual Interpretability of Retrieval}
\label{subsec:visual-interpretability-of-retrieval}

\begin{figure}[htbp]
    \centering
    \includegraphics[width=\linewidth]{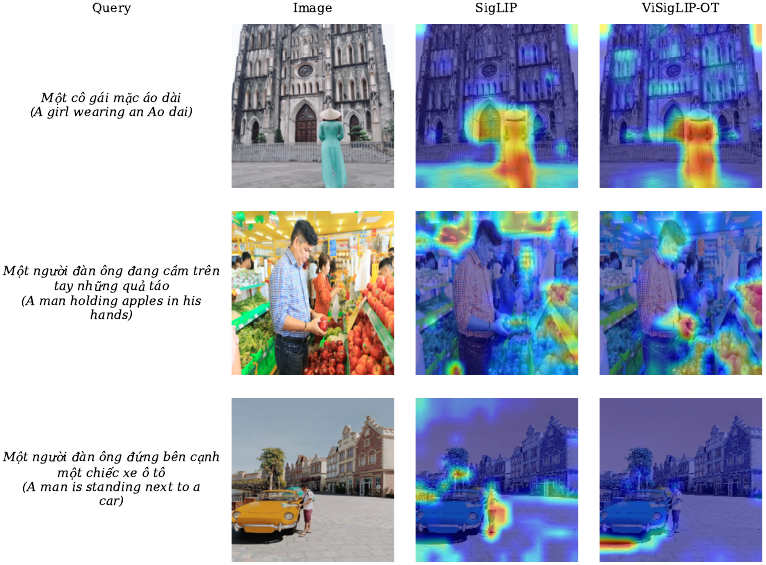}
    \caption{GradCAM visualization comparing the baseline SigLIP and the proposed ViSigLIP-OT on the UIT-OpenViIC test set. Each row shows the original image alongside the GradCAM heatmaps from both models for a given Vietnamese text query. In the first two rows, ViSigLIP-OT focuses more precisely on the query-relevant objects (the girl wearing an \textit{Ao dai} and the man holding apples in his hands), while SigLIP spreads activations over background regions. In the third row, SigLIP correctly attends to the man standing next to a car, whereas ViSigLIP-OT highlights irrelevant background areas.}
    \label{fig:viz-gradcam}
\end{figure}

To further assess how the SIGROT objective influences the model's visual attention, GradCAM~\cite{selvaraju2017grad} is applied to visualize the regions of the input image that contribute most to the image-text similarity score. Figure~\ref{fig:viz-gradcam} presents GradCAM heatmaps for the baseline SigLIP and the proposed ViSigLIP-OT on selected examples from the UIT-OpenViIC test set. In the first two rows, ViSigLIP-OT produces activation maps that are more concentrated on the objects mentioned in the query: the girl wearing an \textit{Ao dai} (row~1) and the man holding apples in his hands (row~2), whereas SigLIP tends to spread activations over broader background regions. This suggests that the SIGROT loss encourages the model to attend to semantically relevant objects rather than contextual surroundings. However, ViSigLIP-OT does not uniformly outperform SigLIP. In the third row, SigLIP correctly highlights the man standing next to a car, while ViSigLIP-OT attends to less relevant background areas. Additional GradCAM visualizations are provided in ~\ref{app:additional-visual-interpretability-of-retrieval}.

\section{Ablation study}\label{sec:ablation_study}
This section presents ablation studies to examine the sensitivity of ViCLIP-OT to key design choices, including partial fine-tuning of the image encoder (Section~\ref{ablation:unfreeze_image_encoder}), the hybrid loss weight $\lambda$ (Section~\ref{ablation:clip_loss_lambda}), and the similarity graph combination strategy (Section~\ref{ablation:combine_method_similarity_graph}). All ablation experiments are conducted on the UIT-OpenViIC validation set.



\subsection{Effect of Partial Fine-Tuning of the Image Encoder}\label{ablation:unfreeze_image_encoder}

\begin{figure}[tbp]
    \centering
    \begin{subfigure}[b]{0.48\linewidth}
        \centering
        \begin{tikzpicture}
            \begin{axis}[
                width=\linewidth,
                height=0.7\linewidth,
                xlabel={\texttt{\#} Last Unfrozen Groups},
                ylabel={Average Recall@K (\%)},
                xmin=0, xmax=15,
                ymin=57, ymax=73,
                xtick={0, 1, 3, 5, 7, 9, 11, 13},
                ytick={57, 61, 65, 69},
                grid=major,
                legend pos=south east,
            ]
            \addplot[
                color=blue,
                mark=square*,
                thick,
            ] coordinates {
                (0, 58.68) (1, 58.96) (2, 65.72) (3, 67.80) (5, 69.27)
                (7, 69.38) (9, 69.60) (11, 69.29) (13, 69.62) (14, 69.04)
            };
            \addlegendentry{Avg. Recall@K}
            \draw [dashed, red] (axis cs:13,57) -- (axis cs:13,71) node [pos=0.95, above, font=\footnotesize] {\textbf{69.62\%}};
            \end{axis}
        \end{tikzpicture}
        \caption{Effect of unfrozen groups.}
        \label{fig:avg_recall_vs_unfrozen_groups}
    \end{subfigure}
    \hfill
    \begin{subfigure}[b]{0.48\linewidth}
        \centering
        \begin{tikzpicture}
            \begin{axis}[
                width=\linewidth,
                height=0.7\linewidth,
                xlabel={$\lambda$},
                xmin=0, xmax=0.55,
                ymin=55, ymax=72,
                xtick={0, 0.1, 0.2, 0.3, 0.5},
                ytick={55, 59, 63, 67, 71},
                grid=major,
                legend pos=south east,
            ]
            \addplot[color=blue, mark=square*, thick] coordinates {
                (0.0, 56.79) (0.1, 69.04) (0.2, 69.20) (0.3, 68.11) (0.5, 67.07)
            };
            \addlegendentry{ViCLIP-OT}
            \addplot[color=orange, mark=triangle*, thick] coordinates {
                (0.0, 56.79) (0.1, 70.76) (0.15, 70.27) (0.2, 70.03) (0.3, 69.45) (0.5, 68.87)
            };
            \addlegendentry{ViSigLIP-OT}
            \draw [dashed, blue] (axis cs:0.2,55) -- (axis cs:0.2,69.20);
            \draw [dashed, orange] (axis cs:0.1,55) -- (axis cs:0.1,70.76);
            \end{axis}
        \end{tikzpicture}
        \caption{Effect of $\lambda$.}
        \label{fig:avg_recall_vs_lambda}
    \end{subfigure}
    \caption{Effect of (a) the number of last unfrozen groups in the image encoder using ViCLIP-OT, and (b) the hybrid loss weight $\lambda$ where $\lambda=0$ corresponds to SIGROT only. Peak performance occurs at 13 unfrozen groups (69.62\%) and $\lambda=0.2$ for ViCLIP-OT, $\lambda=0.1$ for ViSigLIP-OT.}
    \label{fig:ablation_studies}
\end{figure}
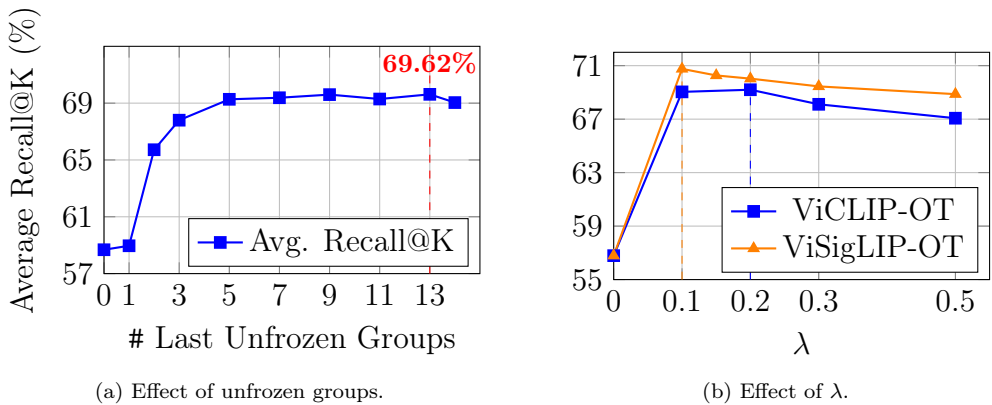

To investigate the impact of fine-tuning different portions of the image encoder, an ablation study was conducted where varying numbers of the last transformer groups in the DINOv3 image encoder were unfrozen during training. The results, presented in Figure~\ref{fig:avg_recall_vs_unfrozen_groups} and Table~\ref{tab:unlock_groups} (\ref{app:detailed-ablation-results}), show that partial unfreezing leads to significant improvements over the fully frozen setting. Unfreezing the last 13 groups yielded the best average R@K of 69.62\%, while unfreezing too many layers (e.g., all 14 groups) caused a slight performance drop, likely due to overfitting or destabilization of pretrained features. The largest gain was observed when unfreezing just the last 2 groups, with an improvement of nearly 7 percentage points, suggesting that even modest adaptation of higher-level visual features is highly beneficial. This is likely attributable to the domain gap between the general-purpose data used to pretrain DINOv3 and UIT-OpenViIC, which contains diverse Vietnamese-related images collected through web searches. This observation is also consistent with prior findings~\cite{zhai2021lit}, where training the image tower with a smaller learning rate improved retrieval performance compared to the Locked-image Text tuning paradigm~\cite{zhai2021lit}, in which only the text tower is tuned to align its embeddings to a frozen image tower. For simplicity and reproducibility, all image tower layers are fine-tuned in the main experiments.

\subsection{Effect of Hybrid Loss Weight $\lambda$}\label{ablation:clip_loss_lambda}

Sensitivity analysis of the hybrid loss weight $\lambda$ in Eq.~\eqref{eq:clip_sigrot_loss} and Eq.~\eqref{eq:siglip_sigrot_loss} was performed to assess its impact on retrieval performance. As shown in Figure~\ref{fig:avg_recall_vs_lambda} and Tables~\ref{tab:clip_loss_lambda} and~\ref{tab:siglip_loss_lambda} (\ref{app:detailed-ablation-results}), varying $\lambda$ from 0.0 to 0.5 revealed that incorporating the contrastive loss (CLIP or SigLIP) alongside the SIGROT loss significantly enhances retrieval metrics compared to using SIGROT alone ($\lambda=0.0$). For ViCLIP-OT, the optimal performance was achieved at $\lambda=0.2$, yielding an average R@K of 69.20\%. For ViSigLIP-OT, the best result was observed at $\lambda=0.1$, achieving an average R@K of 70.76\%. In both cases, increasing $\lambda$ beyond the optimal value led to a gradual decline in performance, indicating that while the contrastive loss is beneficial, excessive weighting can overshadow the advantages provided by the SIGROT regularization. These findings underscore the importance of balancing the contributions of both loss components to maximize cross-modal retrieval effectiveness. In the main experiments, $\lambda = 0.1$ is used for both objectives.

\subsection{Effect of Similarity Graph Combination Strategy}\label{ablation:combine_method_similarity_graph}

To evaluate the impact of different combination strategies for constructing the similarity graph used in SIGROT, an ablation study was conducted. Various methods for combining the text--text, image--image, and cross-modality similarity matrices were tested, including arithmetic mean, harmonic mean, using only caption similarities, using only image similarities, and cross-modality similarities.

Eq.~\eqref{eq:cross_modal_similarity_graph} presents the \emph{cross-modality} combination strategy ($G_{\text{cross}}$) used in the main experiments. The \emph{arithmetic mean} strategy combines the text--text and image--image similarity graphs as follows:
\begin{equation}\label{eq:arithmetic_mean_similarity_graph}
G_{\text{arith}} = \frac{1}{2}\left(G_{\texttext} +G_{\imageimage}\right).
\end{equation}

Whereas for \emph{harmonic mean}, the similarity graph is computed as:
\begin{equation}\label{eq:harmonic_mean_similarity_graph}
G_{\text{harm}} = 2 \left( \frac{1}{G_{\texttext}} + \frac{1}{G_{\imageimage}} \right)^{-1} = \frac{2 G_{\texttext}G_{\imageimage}}{G_{\texttext} +G_{\imageimage}}.
\end{equation}

Similarly, the \emph{caption only} and \emph{image only} strategies directly use $G_{\texttext}$ and $G_{\imageimage}$ as the similarity graph, respectively.

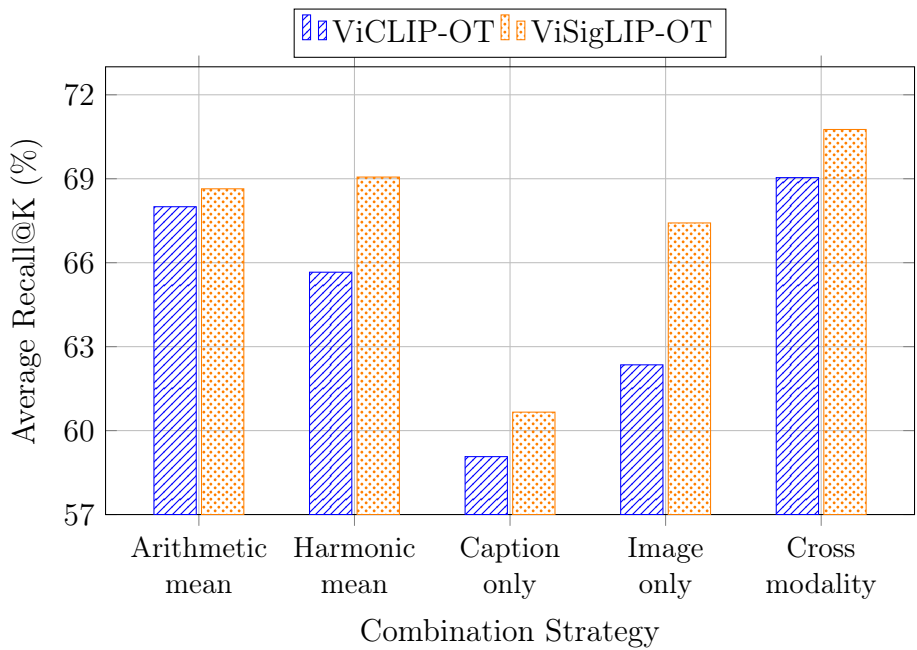
\begin{figure}[tbp]
    \centering
    \begin{tikzpicture}
        \begin{axis}[
            ybar,
            bar width=16pt,
            width=0.9\linewidth,
            height=0.55\linewidth,
            xlabel={Combination Strategy},
            ylabel={Average Recall@K (\%)},
            symbolic x coords={Arithmetic mean, Harmonic mean, Caption only, Image only, Cross modality},
            xticklabels={Arithmetic\\mean, Harmonic\\mean, Caption\\only, Image\\only, Cross\\modality},
            x tick label style={align=center, font=\small},
            xtick=data,
            ymin=57, ymax=73,
            ytick={57, 60, 63, 66, 69, 72},
            grid=major,
            legend style={at={(0.5,1.02)}, anchor=south, legend columns=2},
            enlarge x limits=0.15,
        ]
        \addplot[
            draw=blue!100!black,
            pattern=north east lines,
            pattern color=blue!100!black,
        ] coordinates {
            (Arithmetic mean, 68.00)
            (Harmonic mean, 65.66)
            (Caption only, 59.07)
            (Image only, 62.35)
            (Cross modality, 69.04)
        };
        \addplot[
            draw=orange!100!black,
            pattern=crosshatch dots,
            pattern color=orange!100!black,
        ] coordinates {
            (Arithmetic mean, 68.64)
            (Harmonic mean, 69.06)
            (Caption only, 60.66)
            (Image only, 67.42)
            (Cross modality, 70.76)
        };
        \legend{ViCLIP-OT, ViSigLIP-OT}
        \end{axis}
    \end{tikzpicture}
    \caption{Average Recall@K for different similarity graph combination strategies. The cross-modality approach achieves the highest performance for both loss configurations.}
    \label{fig:avg_recall_vs_combination_strategy}
\end{figure}

The results, summarized in Figure~\ref{fig:avg_recall_vs_combination_strategy} and Table~\ref{tab:combine_method_similarity_graph} (\ref{app:detailed-ablation-results}), indicate that the cross-modality combination strategy yields the best overall retrieval performance, achieving an average R@K of 69.04\% for ViCLIP-OT and 70.76\% for ViSigLIP-OT. This approach effectively captures both intra-modal and inter-modal similarities, enhancing the model's ability to align images and texts. In contrast, single-modality strategies (caption only or image only), which rely on similarities within a single modality, resulted in the lowest performance. Dual-modality strategies (arithmetic mean and harmonic mean), which aggregate similarities from both modalities, improve over single-modality ones but still fall short of the cross-modality strategy. These results underscore the importance of integrating inter-modal information in the similarity graph construction.

Combining the insights from these ablation studies, it is evident that careful selection of model components and hyperparameters significantly influences the effectiveness of ViCLIP-OT in cross-modal retrieval tasks. The findings highlight the importance of fine-tuning strategies, loss function balancing, and similarity graph construction in optimizing model performance.

\section{Conclusion}\label{sec:conclusion}

This study presents ViCLIP-OT, a foundation vision--language model tailored for Vietnamese image--text retrieval. By combining contrastive learning with a similarity-graph regularized optimal transport objective, the proposed approach enhances both instance-level alignment and distribution-level structural consistency across modalities.

Experimental evaluations on UIT-OpenViIC, KTVIC, and Crossmodal-3600 demonstrate consistent improvements over strong CLIP and SigLIP baselines in both in-domain and zero-shot scenarios. The hybrid objective not only increases R@K performance but also reduces the modality gap between image and text embeddings, resulting in more coherent shared latent space. Ablation analyses confirm the importance of partial fine-tuning, balanced hybrid loss weighting, and cross-modality graph construction.

The findings suggest that optimal transport-based structural regularization is a promising direction for improving cross-modal retrieval under limited-resource conditions. Future work may investigate large-scale pretraining strategies, end-to-end similarity graph learning, and extensions to other multimodal expert systems such as visual question answering and multimodal reasoning frameworks.


\section*{Acknowledgments}

The authors would like to thank CICT, Can Tho University (CTU) for providing computational resources that partially supported the training of the models used in this study.

\section*{Data Availability}
To support reproducibility and future research, the official implementation and pretrained models are publicly available:

\begin{itemize}
\item Code repository:
\href{https://github.com/minhnguyent546/ViCLIP-OT}{{\color{black}\faGithub}\ minhnguyent546/ViCLIP-OT}
\item Pretrained models:
\href{https://huggingface.co/collections/minhnguyent546/viclip-ot}{\includegraphics[height=1em, trim=20 30 20 20, clip]{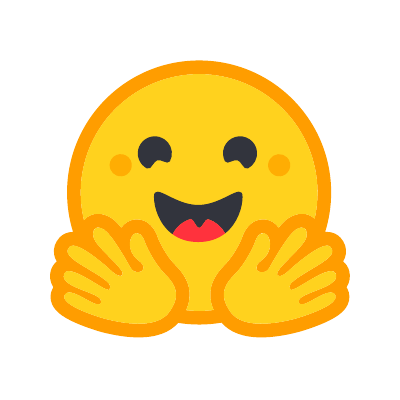}\ minhnguyent546/viclip-ot}
\end{itemize}

\bibliographystyle{plain}
\bibliography{refs}

@misc{simeoni2025dinov3,
  title         = {{DINOv3}},
  author        = {Sim{\'e}oni, Oriane and Vo, Huy V. and Seitzer, Maximilian and Baldassarre, Federico and Oquab, Maxime and Jose, Cijo and Khalidov, Vasil and Szafraniec, Marc and Yi, Seungeun and Ramamonjisoa, Micha{\"e}l and Massa, Francisco and Haziza, Daniel and Wehrstedt, Luca and Wang, Jianyuan and Darcet, Timoth{\'e}e and Moutakanni, Th{\'e}o and Sentana, Leonel and Roberts, Claire and Vedaldi, Andrea and Tolan, Jamie and Brandt, John and Couprie, Camille and Mairal, Julien and J{\'e}gou, Herv{\'e} and Labatut, Patrick and Bojanowski, Piotr},
  year          = {2025},
  eprint        = {2508.10104},
  archiveprefix = {arXiv},
  primaryclass  = {cs.CV},
  url           = {https://arxiv.org/abs/2508.10104}
}

@article{khamis24OT,
  title     = {Scalable Optimal Transport Methods in Machine Learning: A Contemporary Survey},
  author    = {Abdelwahed Khamis and Russell Tsuchida and Mohamed Tarek and Vivien Rolland and Lars Petersson},
  year      = {2024},
  journal   = {IEEE Transactions on Pattern Analysis and Machine Intelligence},
  publisher = {IEEE}
}

@article{cuturi2013sinkhorndistanceslightspeedcomputation,
  title={Sinkhorn distances: Lightspeed computation of optimal transport},
  author={Cuturi, Marco},
  journal={Advances in neural information processing systems},
  volume={26},
  year={2013}
}

@article{BUI2026117430,
  title    = {UIT-OpenViIC: An open-domain benchmark for evaluating image captioning in Vietnamese},
  journal  = {Signal Processing: Image Communication},
  volume   = {140},
  pages    = {117430},
  year     = {2026},
  issn     = {0923-5965},
  doi      = {https://doi.org/10.1016/j.image.2025.117430},
  url      = {https://www.sciencedirect.com/science/article/pii/S0923596525001766},
  author   = {Doanh C. Bui and Nghia Hieu Nguyen and Khang Nguyen}
}

@inproceedings{sobal2024mathbbxsamplecontrastivelossimproving,
 author = {Sobal, Vlad and Ibrahim, Mark and Balestriero, Randall and Cabannes, Vivien and Bouchacourt, Diane and Astolfi, Pietro and Cho, Kyunghyun and LeCun, Yann},
 booktitle = {International Conference on Learning Representations},
 editor = {Y. Yue and A. Garg and N. Peng and F. Sha and R. Yu},
 pages = {70564--70585},
 title = {$\mathbb{X}$-Sample Contrastive Loss: Improving Contrastive Learning with Sample Similarity Graphs},
 url = {https://proceedings.iclr.cc/paper_files/paper/2025/file/afe37ac3ce109cd33a23a6b3ed0cfc21-Paper-Conference.pdf},
 volume = {2025},
 year = {2025}
}

@inproceedings{pmlr-v139-radford21a,
  title     = {Learning Transferable Visual Models From Natural Language Supervision},
  author    = {Radford, Alec and Kim, Jong Wook and Hallacy, Chris and Ramesh, Aditya and Goh, Gabriel and Agarwal, Sandhini and Sastry, Girish and Askell, Amanda and Mishkin, Pamela and Clark, Jack and Krueger, Gretchen and Sutskever, Ilya},
  booktitle = {Proceedings of the 38th International Conference on Machine Learning},
  pages     = {8748--8763},
  year      = {2021},
  editor    = {Meila, Marina and Zhang, Tong},
  volume    = {139},
  series    = {Proceedings of Machine Learning Research},
  month     = {18--24 Jul},
  publisher = {PMLR},
  url       = {https://proceedings.mlr.press/v139/radford21a.html}
}

@inproceedings{jia2021scaling,
  title        = {Scaling up visual and vision-language representation learning with noisy text supervision},
  author       = {Jia, Chao and Yang, Yinfei and Xia, Ye and Chen, Yi-Ting and Parekh, Zarana and Pham, Hieu and Le, Quoc and Sung, Yun-Hsuan and Li, Zhen and Duerig, Tom},
  booktitle    = {International conference on machine learning},
  pages        = {4904--4916},
  year         = {2021},
  organization = {PMLR}
}

@inproceedings{10.1007/978-3-030-63007-2_57,
  author    = {Lam, Quan Hoang
               and Le, Quang Duy
               and Nguyen, Van Kiet
               and Nguyen, Ngan Luu-Thuy},
  editor    = {Nguyen, Ngoc Thanh
               and Hoang, Bao Hung
               and Huynh, Cong Phap
               and Hwang, Dosam
               and Trawi{\'{n}}ski, Bogdan
               and Vossen, Gottfried},
  title     = {UIT-ViIC: A Dataset for the First Evaluation on Vietnamese Image Captioning},
  booktitle = {Computational Collective Intelligence},
  year      = {2020},
  publisher = {Springer International Publishing},
  address   = {Cham},
  pages     = {730--742},
  isbn      = {978-3-030-63007-2}
}

@misc{pham2024ktvicvietnameseimagecaptioning,
  title         = {KTVIC: A Vietnamese Image Captioning Dataset on the Life Domain},
  author        = {Anh-Cuong Pham and Van-Quang Nguyen and Thi-Hong Vuong and Quang-Thuy Ha},
  year          = {2024},
  eprint        = {2401.08100},
  archiveprefix = {arXiv},
  primaryclass  = {cs.CV},
  url           = {https://arxiv.org/abs/2401.08100}
}

@inproceedings{shi2023understanding,
  title        = {Understanding and generalizing contrastive learning from the inverse optimal transport perspective},
  author       = {Shi, Liangliang and Zhang, Gu and Zhen, Haoyu and Fan, Jintao and Yan, Junchi},
  booktitle    = {International conference on machine learning},
  pages        = {31408--31421},
  year         = {2023},
  organization = {PMLR}
}

@article{Luu_Thuy_Nguyen_2023,
  title     = {EVJVQA CHALLENGE: MULTILINGUAL VISUAL QUESTION ANSWERING},
  issn      = {1813-9663},
  url       = {http://dx.doi.org/10.15625/1813-9663/18157},
  doi       = {10.15625/1813-9663/18157},
  journal   = {Journal of Computer Science and Cybernetics},
  publisher = {Publishing House for Science and Technology, Vietnam Academy of Science and Technology (Publications)},
  author    = {Luu-Thuy Nguyen, Ngan and Nghia Hieu Nguyen and T.D. Vo, Duong and Tran, Khanh Quoc and Nguyen, Kiet Van},
  year      = {2023},
  month     = sep,
  pages     = {237–258}
}

@inproceedings{zhai2022litzeroshottransferlockedimage,
  title={Lit: Zero-shot transfer with locked-image text tuning},
  author={Zhai, Xiaohua and Wang, Xiao and Mustafa, Basil and Steiner, Andreas and Keysers, Daniel and Kolesnikov, Alexander and Beyer, Lucas},
  booktitle={Proceedings of the IEEE/CVF conference on computer vision and pattern recognition},
  pages={18123--18133},
  year={2022}
}

@article{10.3389/fcomp.2024.1473457,
  author   = {Shen, Xiaorong  and Huang, Maowei  and Hu, Zheng  and Cai, Shimin  and Zhou, Tao },
  title    = {Multimodal Fake News Detection with Contrastive Learning and Optimal Transport},
  journal  = {Frontiers in Computer Science},
  volume   = {Volume 6 - 2024},
  year     = {2024},
  url      = {https://www.frontiersin.org/journals/computer-science/articles/10.3389/fcomp.2024.1473457},
  doi      = {10.3389/fcomp.2024.1473457},
  issn     = {2624-9898}
}

@article{montesuma2024recent,
  title     = {Recent advances in optimal transport for machine learning},
  author    = {Montesuma, Eduardo Fernandes and Mboula, Fred Maurice Ngole and Souloumiac, Antoine},
  journal   = {IEEE Transactions on Pattern Analysis and Machine Intelligence},
  year      = {2024},
  publisher = {IEEE}
}

@misc{csizmadia2025distillclipdclipenhancing,
  title         = {Distill CLIP (DCLIP): Enhancing Image-Text Retrieval via Cross-Modal Transformer Distillation},
  author        = {Daniel Csizmadia and Andrei Codreanu and Victor Sim and Vighnesh Prabhu and Michael Lu and Kevin Zhu and Sean O'Brien and Vasu Sharma},
  year          = {2025},
  eprint        = {2505.21549},
  archiveprefix = {arXiv},
  primaryclass  = {cs.CV},
  url           = {https://arxiv.org/abs/2505.21549}
}

@misc{ambrosio2002optimaltransportmapsmongekantorovich,
  title         = {Optimal transport maps in Monge-Kantorovich problem},
  author        = {Luigi Ambrosio},
  year          = {2002},
  note          = {Lecture Notes in ICM, vol. III, pp. 131--140},
}

@InProceedings{schall2024optimizingclipmodelsimage,
author="Schall, Konstantin
and Barthel, Kai Uwe
and Hezel, Nico
and Jung, Klaus",
editor="Ch{\'a}vez, Edgar
and Kimia, Benjamin
and Loko{\v{c}}, Jakub
and Patella, Marco
and Sedmidubsky, Jan",
title="Optimizing CLIP Models for Image Retrieval with Maintained Joint-Embedding Alignment",
booktitle="Similarity Search and Applications",
year="2025",
publisher="Springer Nature Switzerland",
address="Cham",
pages="97--110",
isbn="978-3-031-75823-2"
}

@article{YANG2024110273,
  title    = {Continual learning for cross-modal image-text retrieval based on domain-selective attention},
  journal  = {Pattern Recognition},
  volume   = {149},
  pages    = {110273},
  year     = {2024},
  issn     = {0031-3203},
  doi      = {https://doi.org/10.1016/j.patcog.2024.110273},
  url      = {https://www.sciencedirect.com/science/article/pii/S0031320324000244},
  author   = {Rui Yang and Shuang Wang and Yu Gu and Jihui Wang and Yingzhi Sun and Huan Zhang and Yu Liao and Licheng Jiao}
}

@misc{cao2022imagetextretrievalsurveyrecent,
  title         = {Image-text Retrieval: A Survey on Recent Research and Development},
  author        = {Min Cao and Shiping Li and Juntao Li and Liqiang Nie and Min Zhang},
  year          = {2022},
  eprint        = {2203.14713},
  archiveprefix = {arXiv},
  primaryclass  = {cs.IR},
  url           = {https://arxiv.org/abs/2203.14713}
}

@misc{chen2020simpleframeworkcontrastivelearning,
  title         = {A Simple Framework for Contrastive Learning of Visual Representations},
  author        = {Ting Chen and Simon Kornblith and Mohammad Norouzi and Geoffrey Hinton},
  year          = {2020},
  eprint        = {2002.05709},
  archiveprefix = {arXiv},
  primaryclass  = {cs.LG},
  url           = {https://arxiv.org/abs/2002.05709}
}

@inproceedings{thapliyal2022crossmodal,
  title     = {Crossmodal-3600: A massively multilingual multimodal evaluation dataset},
  author    = {Thapliyal, Ashish V and Tuset, Jordi Pont and Chen, Xi and Soricut, Radu},
  booktitle = {Proceedings of the 2022 Conference on Empirical Methods in Natural Language Processing},
  pages     = {715--729},
  year      = {2022}
}

@article{hu2024comprehensive,
  title     = {A comprehensive survey on contrastive learning},
  author    = {Hu, Haigen and Wang, Xiaoyuan and Zhang, Yan and Chen, Qi and Guan, Qiu},
  journal   = {Neurocomputing},
  volume    = {610},
  pages     = {128645},
  year      = {2024},
  publisher = {Elsevier}
}

@article{qwen3vlembedding,
  title   = {Qwen3-VL-Embedding and Qwen3-VL-Reranker: A Unified Framework for State-of-the-Art Multimodal Retrieval and Ranking},
  author  = {Li, Mingxin and Zhang, Yanzhao and Long, Dingkun and Chen Keqin and Song, Sibo and Bai, Shuai and Yang, Zhibo and Xie, Pengjun and Yang, An and Liu, Dayiheng and Zhou, Jingren and Lin, Junyang},
  journal = {arXiv preprint arXiv:2601.04720},
  year    = {2026}
}

@inproceedings{zhai2023sigmoid,
  title     = {Sigmoid loss for language image pre-training},
  author    = {Zhai, Xiaohua and Mustafa, Basil and Kolesnikov, Alexander and Beyer, Lucas},
  booktitle = {Proceedings of the IEEE/CVF international conference on computer vision},
  pages     = {11975--11986},
  year      = {2023}
}

@inproceedings{zhai2021lit,
  title     = {Lit: Zero-shot transfer with locked-image text tuning. 2022 IEEE},
  author    = {Zhai, Xiaohua and Wang, Xiao and Mustafa, Basil and Steiner, Andreas and Keysers, Daniel and Kolesnikov, Alexander and Beyer, Lucas},
  booktitle = {CVF Conference on Computer Vision and Pattern Recognition (CVPR)},
  pages     = {18102--18112},
  year      = {2021}
}

@inproceedings{wu2018unsupervised,
  title     = {Unsupervised feature learning via non-parametric instance discrimination},
  author    = {Wu, Zhirong and Xiong, Yuanjun and Yu, Stella X and Lin, Dahua},
  booktitle = {Proceedings of the IEEE conference on computer vision and pattern recognition},
  pages     = {3733--3742},
  year      = {2018}
}

@article{mcinnes2018umap,
  title   = {Umap: Uniform manifold approximation and projection for dimension reduction},
  author  = {McInnes, Leland and Healy, John and Melville, James},
  journal = {arXiv preprint arXiv:1802.03426},
  year    = {2018}
}

@inproceedings{gao2021scaling,
     title={Scaling Deep Contrastive Learning Batch Size under Memory Limited Setup},
     author={Luyu Gao and Yunyi Zhang and Jiawei Han and Jamie Callan},
     booktitle ={Proceedings of the 6th Workshop on Representation Learning for NLP},
     year={2021},
}

@article{liang2022mind,
  title   = {Mind the gap: Understanding the modality gap in multi-modal contrastive representation learning},
  author  = {Liang, Victor Weixin and Zhang, Yuhui and Kwon, Yongchan and Yeung, Serena and Zou, James Y},
  journal = {Advances in Neural Information Processing Systems},
  volume  = {35},
  pages   = {17612--17625},
  year    = {2022}
}

@inproceedings{eslami2025mitigate,
  title     = {Mitigate the gap: Improving cross-modal alignment in CLIP},
  author    = {Eslami, Sedigheh and de Melo, Gerard},
  booktitle = {The Thirteenth International Conference on Learning Representations},
  year      = {2025}
}

@article{goel2022cyclip,
  title   = {Cyclip: Cyclic contrastive language-image pretraining},
  author  = {Goel, Shashank and Bansal, Hritik and Bhatia, Sumit and Rossi, Ryan and Vinay, Vishwa and Grover, Aditya},
  journal = {Advances in Neural Information Processing Systems},
  volume  = {35},
  pages   = {6704--6719},
  year    = {2022}
}

@inproceedings{reimers-2019-sentence-bert,
  title = "Sentence-BERT: Sentence Embeddings using Siamese BERT-Networks",
  author = "Reimers, Nils and Gurevych, Iryna",
  booktitle = "Proceedings of the 2019 Conference on Empirical Methods in Natural Language Processing",
  month = "11",
  year = "2019",
  publisher = "Association for Computational Linguistics",
  url = "https://arxiv.org/abs/1908.10084",
}

@inproceedings{gunther2025jina,
  title={jina-embeddings-v4: Universal embeddings for multimodal multilingual retrieval},
  author={G{\"u}nther, Michael and Sturua, Saba and Akram, Mohammad Kalim and Mohr, Isabelle and Ungureanu, Andrei and Wang, Bo and Eslami, Sedigheh and Martens, Scott and Werk, Maximilian and Wang, Nan and others},
  booktitle={Proceedings of the 5th Workshop on Multilingual Representation Learning (MRL 2025)},
  pages={531--550},
  year={2025}
}

@article{pizzi2022self,
  title={A Self-Supervised Descriptor for Image Copy Detection},
  author={Pizzi, Ed and Roy, Sreya Dutta and Ravindra, Sugosh Nagavara and Goyal, Priya and Douze, Matthijs},
  journal={Proc. CVPR},
  year={2022}
}

@misc{koukounas2024jinaclipv2multilingualmultimodalembeddings,
      title={jina-clip-v2: Multilingual Multimodal Embeddings for Text and Images}, 
      author={Andreas Koukounas and Georgios Mastrapas and Bo Wang and Mohammad Kalim Akram and Sedigheh Eslami and Michael Günther and Isabelle Mohr and Saba Sturua and Scott Martens and Nan Wang and Han Xiao},
      year={2024},
      eprint={2412.08802},
      archivePrefix={arXiv},
      primaryClass={cs.CL},
      url={https://arxiv.org/abs/2412.08802}, 
}

@article{loshchilov2017decoupled,
  title={Decoupled weight decay regularization},
  author={Loshchilov, Ilya and Hutter, Frank},
  journal={arXiv preprint arXiv:1711.05101},
  year={2017}
}

@inproceedings{selvaraju2017grad,
  title={Grad-cam: Visual explanations from deep networks via gradient-based localization},
  author={Selvaraju, Ramprasaath R and Cogswell, Michael and Das, Abhishek and Vedantam, Ramakrishna and Parikh, Devi and Batra, Dhruv},
  booktitle={Proceedings of the IEEE international conference on computer vision},
  pages={618--626},
  year={2017}
}

@inproceedings{shi2023towards,
  title     = {Towards understanding the modality gap in {CLIP}},
  author    = {Peiyang Shi and Michael C. Welle and M{\r{a}}rten Bj{\"o}rkman and Danica Kragic},
  booktitle = {ICLR 2023 Workshop on Multimodal Representation Learning: Perks and Pitfalls},
  year      = {2023},
  url       = {https://openreview.net/forum?id=8W3KGzw7fNI}
}

@InProceedings{yaras2024explaining,
  title = 	 {Explaining and Mitigating the Modality Gap in Contrastive Multimodal Learning},
  author =       {Yaras, Can and Chen, Siyi and Wang, Peng and Qu, Qing},
  booktitle = 	 {Conference on Parsimony and Learning},
  pages = 	 {1365--1387},
  year = 	 {2025},
  editor = 	 {Chen, Beidi and Liu, Shijia and Pilanci, Mert and Su, Weijie and Sulam, Jeremias and Wang, Yuxiang and Zhu, Zhihui},
  volume = 	 {280},
  series = 	 {Proceedings of Machine Learning Research},
  month = 	 {24--27 Mar},
  publisher =    {PMLR},
  url = 	 {https://proceedings.mlr.press/v280/yaras25a.html}
}

@inproceedings{sofer2025pull,
  title     = {Pull It Together: Reducing the Modality Gap in Contrastive Learning},
  author    = {Sofer, Amit and Goldman, Yoav and Chazan, Shlomo E},
  booktitle = {Proc. Interspeech 2025},
  pages     = {196--200},
  year      = {2025}
}

@article{vera2025embeddinggemma,
  title   = {Embeddinggemma: Powerful and lightweight text representations},
  author  = {Vera, Henrique Schechter and Dua, Sahil and Zhang, Biao and Salz, Daniel and Mullins, Ryan and Panyam, Sindhu Raghuram and Smoot, Sara and Naim, Iftekhar and Zou, Joe and Chen, Feiyang and others},
  journal = {arXiv preprint arXiv:2509.20354},
  year    = {2025}
}

@article{chen2024bge,
  title   = {Bge m3-embedding: Multi-lingual, multi-functionality, multi-granularity text embeddings through self-knowledge distillation},
  author  = {Chen, Jianlv and Xiao, Shitao and Zhang, Peitian and Luo, Kun and Lian, Defu and Liu, Zheng},
  journal = {arXiv preprint arXiv:2402.03216},
  volume  = {4},
  number  = {5},
  year    = {2024}
}

@article{muennighoff2022mteb,
  author    = {Muennighoff, Niklas and Tazi, Nouamane and Magne, Loïc and Reimers, Nils},
  title     = {MTEB: Massive Text Embedding Benchmark},
  publisher = {arXiv},
  journal   = {arXiv preprint arXiv:2210.07316},
  year      = {2022},
  url       = {https://arxiv.org/abs/2210.07316},
  doi       = {10.48550/ARXIV.2210.07316}
}

@article{enevoldsen2025mmtebmassivemultilingualtext,
  title     = {MMTEB: Massive Multilingual Text Embedding Benchmark},
  author    = {Kenneth Enevoldsen and Isaac Chung and Imene Kerboua and Márton Kardos and Ashwin Mathur and David Stap and Jay Gala and Wissam Siblini and Dominik Krzemiński and Genta Indra Winata and Saba Sturua and Saiteja Utpala and Mathieu Ciancone and Marion Schaeffer and Gabriel Sequeira and Diganta Misra and Shreeya Dhakal and Jonathan Rystrøm and Roman Solomatin and Ömer Çağatan and Akash Kundu and Martin Bernstorff and Shitao Xiao and Akshita Sukhlecha and Bhavish Pahwa and Rafał Poświata and Kranthi Kiran GV and Shawon Ashraf and Daniel Auras and Björn Plüster and Jan Philipp Harries and Loïc Magne and Isabelle Mohr and Mariya Hendriksen and Dawei Zhu and Hippolyte Gisserot-Boukhlef and Tom Aarsen and Jan Kostkan and Konrad Wojtasik and Taemin Lee and Marek Šuppa and Crystina Zhang and Roberta Rocca and Mohammed Hamdy and Andrianos Michail and John Yang and Manuel Faysse and Aleksei Vatolin and Nandan Thakur and Manan Dey and Dipam Vasani and Pranjal Chitale and Simone Tedeschi and Nguyen Tai and Artem Snegirev and Michael Günther and Mengzhou Xia and Weijia Shi and Xing Han Lù and Jordan Clive and Gayatri Krishnakumar and Anna Maksimova and Silvan Wehrli and Maria Tikhonova and Henil Panchal and Aleksandr Abramov and Malte Ostendorff and Zheng Liu and Simon Clematide and Lester James Miranda and Alena Fenogenova and Guangyu Song and Ruqiya Bin Safi and Wen-Ding Li and Alessia Borghini and Federico Cassano and Hongjin Su and Jimmy Lin and Howard Yen and Lasse Hansen and Sara Hooker and Chenghao Xiao and Vaibhav Adlakha and Orion Weller and Siva Reddy and Niklas Muennighoff},
  publisher = {arXiv},
  journal   = {arXiv preprint arXiv:2502.13595},
  year      = {2025},
  url       = {https://arxiv.org/abs/2502.13595},
  doi       = {10.48550/arXiv.2502.13595}
}

@article{Sinkhorn1967ConcerningNM,
  title={Concerning nonnegative matrices and doubly stochastic matrices},
  author={Richard Sinkhorn and Paul Knopp},
  journal={Pacific Journal of Mathematics},
  year={1967},
  volume={21},
  pages={343-348},
  url={https://api.semanticscholar.org/CorpusID:50329347}
}

@article{frogner2015learning,
  title={Learning with a Wasserstein loss},
  author={Frogner, Charlie and Zhang, Chiyuan and Mobahi, Hossein and Araya, Mauricio and Poggio, Tomaso A},
  journal={Advances in neural information processing systems},
  volume={28},
  year={2015}
}

@inproceedings{karpathy2015deep,
  title={Deep visual-semantic alignments for generating image descriptions},
  author={Karpathy, Andrej and Fei-Fei, Li},
  booktitle={Proceedings of the IEEE conference on computer vision and pattern recognition},
  pages={3128--3137},
  year={2015}
}

@inproceedings{chen2020uniter,
  title={Uniter: Universal image-text representation learning},
  author={Chen, Yen-Chun and Li, Linjie and Yu, Licheng and El Kholy, Ahmed and Ahmed, Faisal and Gan, Zhe and Cheng, Yu and Liu, Jingjing},
  booktitle={European conference on computer vision},
  pages={104--120},
  year={2020},
  organization={Springer}
}

@inproceedings{zhan2025elip,
  title={Elip: Enhanced visual-language foundation models for image retrieval},
  author={Zhan, Guanqi and Liu, Yuanpei and Han, Kai and Xie, Weidi and Zisserman, Andrew},
  booktitle={2025 International Conference on Content-Based Multimedia Indexing (CBMI)},
  pages={1--8},
  year={2025},
  organization={IEEE}
}





\appendix

\section{Dataset Details}
\label{app:dataset-details}

This appendix provides detailed descriptions of the three datasets used in this work. Table~\ref{tab:dataset-summary} summarizes the key statistics of the datasets.

\paragraph{UIT-OpenViIC}
UIT-OpenViIC~\cite{BUI2026117430} is a large-scale open-domain Vietnamese image captioning dataset. The images were primarily crawled from Google and Bing using Vietnamese search keywords. After collection, images that were duplicated or lacked sufficient visual information were filtered out. The final dataset consists of 13,100 images annotated with a total of 61,241 captions. The dataset is split into training, validation, and test sets following an image-level partitioning strategy. Approximately 30\% of the images and their corresponding captions are reserved for validation and testing, with each split accounting for about 15\% of the total images. Images are randomly sampled using a uniform distribution to ensure equal selection probability. As a result, the training set contains 9,088 images with 41,238 captions, the validation set includes 2,011 images with 10,002 captions, and the test set consists of 2,001 images with 10,001 captions.

\paragraph{KTVIC}
KTVIC~\cite{pham2024ktvicvietnameseimagecaptioning} is a Vietnamese image captioning benchmark constructed from images sourced from the UIT-EVJVQA~\cite{Luu_Thuy_Nguyen_2023} dataset. It comprises 4,327 images annotated with a total of 21,635 captions, where each image is associated with five human-written captions. These multiple annotations capture visual content from diverse perspectives, contributing to linguistic richness and semantic diversity. The images mainly depict daily-life activities of Vietnamese people and various locations within Vietnam. Following the original splits of UIT-EVJVQA, the dataset is divided into a training set of 3,769 images and a test set of 558 images. Near-duplicate detection against the UIT-OpenViIC training set reveals a substantial overlap (see~\ref{app:remove-near-duplicate}): 401 out of 558 test images and 2,464 out of 3,769 training images. After deduplication against the UIT-OpenViIC training set, 157 test images and 1,305 training images remain.

\paragraph{Crossmodal-3600}
Crossmodal-3600 (or \crossmodaldataset)~\cite{thapliyal2022crossmodal} is a geographically diverse collection of 3,600 photos tagged with human-generated reference captions in 36 languages. For each language, 100 images were selected from regions where the language is spoken. This ensures diverse visual contexts across the dataset. The photographs are annotated with captions that achieve uniformity in terms of style across all languages, while avoiding annotation artifacts caused by direct translation. Near-duplicate detection against the UIT-OpenViIC training set yields no matches. In this work, Crossmodal-3600 is employed for zero-shot evaluation to assess model generalization. All 3,600 images paired with Vietnamese captions are used, yielding 7,350 image--caption pairs.

\begin{table*}[tbp]
\centering
\caption{Summary statistics of the datasets used in this work. KTVIC (dedup.) denotes the deduplicated version after near-duplicate removal against the UIT-OpenViIC training set.}
\label{tab:dataset-summary}
\resizebox{\textwidth}{!}{%
\begin{tabular}{l l rr rr rr}
\toprule
\multirow{2}{*}{Dataset} & \multirow{2}{*}{Domain} & \multicolumn{2}{c}{Train} & \multicolumn{2}{c}{Validation} & \multicolumn{2}{c}{Test} \\
\cmidrule(lr){3-4} \cmidrule(lr){5-6} \cmidrule(lr){7-8}
& & \texttt{\#} Img. & \texttt{\#} Cap. & \texttt{\#} Img. & \texttt{\#} Cap. & \texttt{\#} Img. & \texttt{\#} Cap. \\
\midrule
UIT-OpenViIC~\cite{BUI2026117430} & Open domain & 9,088 & 41,238 & 2,011 & 10,002 & 2,001 & 10,001 \\
KTVIC~\cite{pham2024ktvicvietnameseimagecaptioning} & Daily activities & 3,769 & 18,845 & - & - & 558 & 2,790 \\
KTVIC (dedup.) & Daily activities & 1,305 & 6,525 & - & - & 157 & 785 \\
\crossmodaldataset{}~\cite{thapliyal2022crossmodal} & Geographically diverse & - & - & - & - & 3,600 & 7,350 \\
\bottomrule
\end{tabular}
}%
\end{table*}

\section{Near-Duplicate Removal from Evaluation Data}
\label{app:remove-near-duplicate}

To prevent train--test contamination and ensure fair evaluation, a near-duplicate detection step is performed on the evaluation datasets against the UIT-OpenViIC training set. Specifically, SSCD (A Self-Supervised Descriptor for Image Copy Detection)~\cite{pizzi2022self} is employed to identify duplicate and near-duplicate images in the evaluation datasets against the UIT-OpenViIC training set. The pretrained \texttt{sscd\_disc\_mixup}\footnote{\url{https://github.com/facebookresearch/sscd-copy-detection}} model is used, and image pairs with a cosine similarity $\geq 0.8$ are considered near-duplicates. The deduplication is applied to both the training and test splits of KTVIC, as well as to Crossmodal-3600. The UIT-OpenViIC training set is kept intact, as the available training data is limited.

Table~\ref{tab:dedup-results} summarizes the results. A substantial overlap is found in KTVIC: 401 out of 558 test images (71.9\%) and 2,464 out of 3,769 training images (65.4\%) are identified as near-duplicates of UIT-OpenViIC training images. After removal, only 157 test images and 1,305 training images remain. Due to the severely reduced test set, zero-shot evaluation on KTVIC is performed on both the deduplicated training and test splits. No near-duplicates are found for Crossmodal-3600.

\begin{table}[tbp]
\centering
\caption{Near-duplicate removal results against the UIT-OpenViIC training set using SSCD with a cosine similarity threshold of 0.8.}
\label{tab:dedup-results}
\resizebox{0.8\textwidth}{!}{%
\begin{tabular}{l rr rr rr}
\toprule
\multirow{2}{*}{Dataset} & \multicolumn{2}{c}{Original} & \multicolumn{2}{c}{Duplicates} & \multicolumn{2}{c}{Remaining} \\
\cmidrule(lr){2-3} \cmidrule(lr){4-5} \cmidrule(lr){6-7}
& \texttt{\#} Img. & \texttt{\#} Cap. & \texttt{\#} Img. & \texttt{\#} Cap. & \texttt{\#} Img. & \texttt{\#} Cap. \\
\midrule
KTVIC-train & 3,769 & 18,845 & 2,464 & 12,320 & 1,305 & 6,525 \\
KTVIC-test & 558 & 2,790 & 401 & 2,005 & 157 & 785 \\
Crossmodal-3600 & 3,600 & 7,350 & 0 & 0 & 3,600 & 7,350 \\
\bottomrule
\end{tabular}
}%
\end{table}





\section{Modality gap in multimodal contrastive representation learning}
\label{app:modality-gap}

The \textit{modality gap}~\cite{liang2022mind} is a known phenomenon in multimodal representation learning, where embeddings from different modalities occupy distinct regions within the shared latent space. Existing studies have investigated this phenomenon from both theoretical and empirical perspectives. \cite{liang2022mind} demonstrate empirically and theoretically that the modality gap arises from a combination of initialization and optimization dynamics, and that contrastive learning objective encourages the existence of the modality gap. \cite{shi2023towards} experimentally examine the influence of different initialization strategies and temperature parameters on the modality gap using toy datasets, though without providing theoretical justification for its emergence under varying conditions. More recent work has also focused on proposing practical solutions. \cite{yaras2024explaining} analyze the gradient flow dynamics during training and identify mismatched data pairs and a learnable temperature parameter as key contributors to the modality gap; they further show empirically that closing the modality gap is particularly beneficial for image--text retrieval, whereas zero-shot and linear probing on image classification are not strongly correlated with the modality gap but rather with the uniformity of the embedding space. \cite{eslami2025mitigate} propose sharing learnable parameters between modality encoders to improve the alignment between text and image embeddings, while \cite{sofer2025pull} introduce a solution combining a modality classifier with a Gradient Reversal Layer (GRL) to reduce the gap in the text--audio domain.
\section{Additional Visual Interpretability of Retrieval}
\label{app:additional-visual-interpretability-of-retrieval}

\begin{figure}[htbp]
    \centering
    \includegraphics[width=\linewidth]{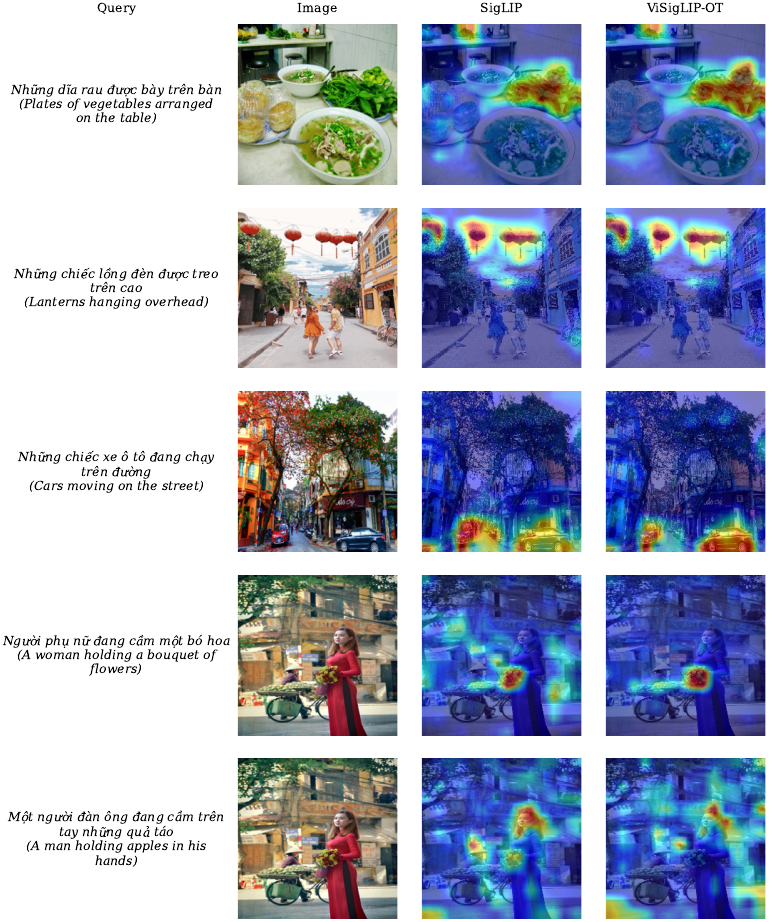}
    \caption{Additional GradCAM visualizations comparing SigLIP and ViSigLIP-OT. When the relevant objects are easy to identify (rows~1--2: plates of vegetables and lanterns), both models produce similar heatmaps. In more complex scenes (rows~3--4: cars on the street and a woman holding flowers), ViSigLIP-OT attends more precisely to the query-relevant objects, while SigLIP distributes attention to background regions. In the last row, the image does not match the query, and ViSigLIP-OT produces a more spread-out heatmap rather than concentrating on specific regions.}
    \label{fig:additional-viz-gradcam}
\end{figure}

Figure~\ref{fig:additional-viz-gradcam} shows additional GradCAM visualizations comparing SigLIP and ViSigLIP-OT on further examples from the UIT-OpenViIC test set.

\section{Qualitative Comparison between SigLIP and ViSigLIP-OT}
\label{app:retrieval-examples}
\begin{figure}[hbtp]
    \centering
    \includegraphics[width=0.865\linewidth]{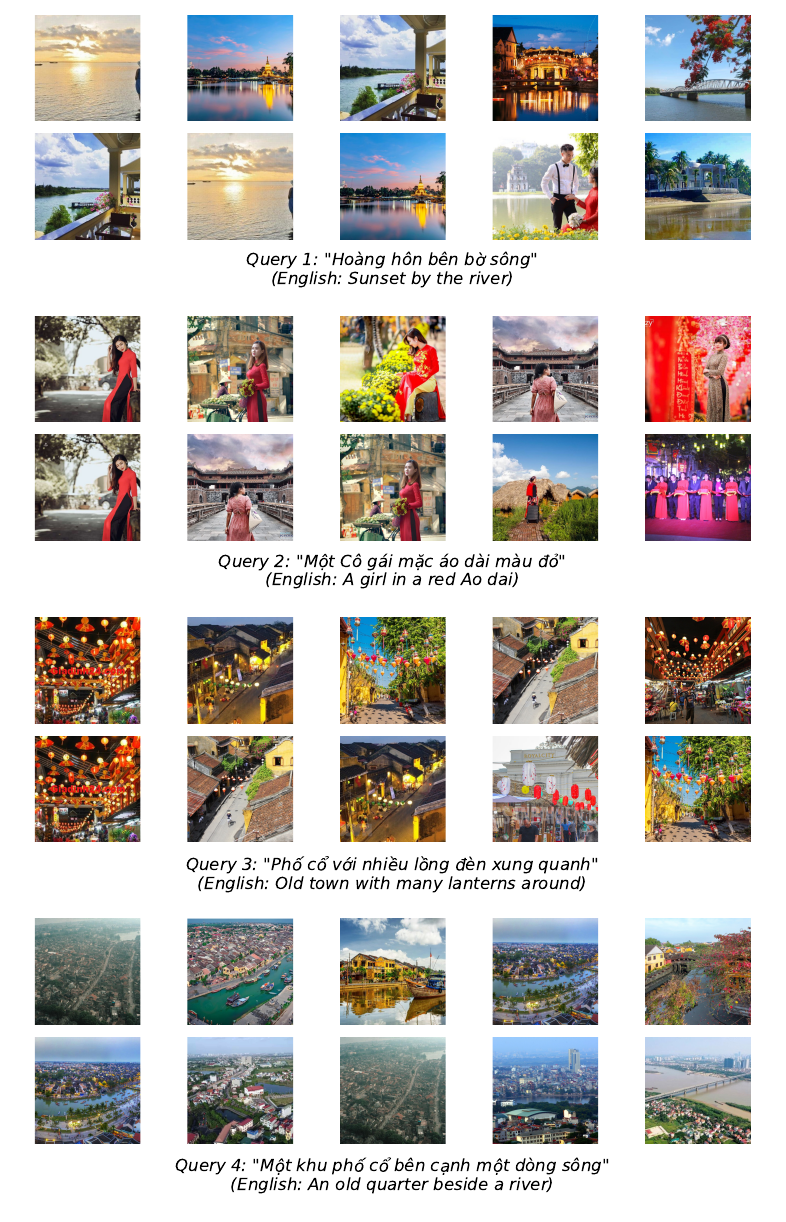}
    \caption{Text-to-image retrieval examples on the UIT-OpenViIC test set. For each query, the first row shows results from ViSigLIP-OT and the second row from SigLIP. Retrieved images are sorted by similarity scores.}
    \label{fig:t2i-retrieval-examples}
\end{figure}

\begin{figure}[hbtp]
    \centering
    \includegraphics[width=0.875\linewidth]{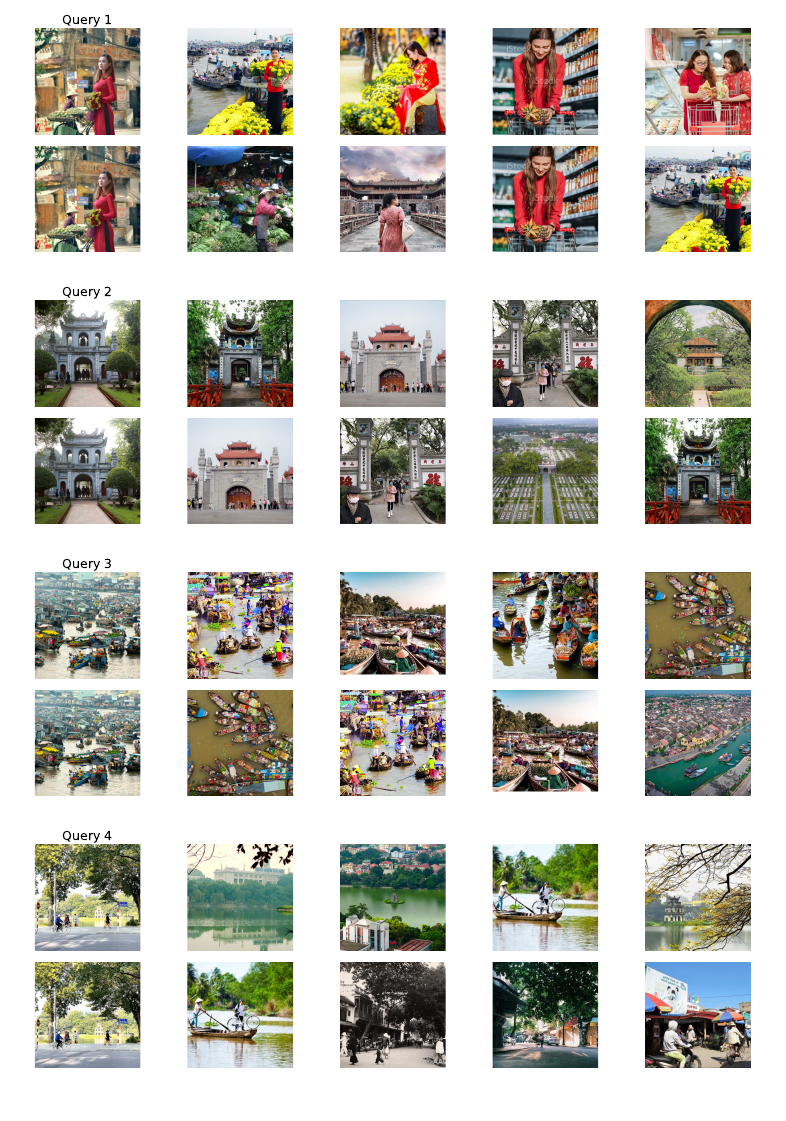}
    \caption{Image-to-image retrieval examples on the UIT-OpenViIC test set. The first image in each row is the query. For each query, the first row shows results from ViSigLIP-OT and the second row from SigLIP. Retrieved images are sorted by similarity scores.}
    \label{fig:i2i-retrieval-examples}
\end{figure}

This appendix provides examples of retrieval results obtained using ViSig\allowbreak{-}CLIP-OT on the UIT-OpenViIC test set. Text-to-image retrieval examples are shown in Figure~\ref{fig:t2i-retrieval-examples} and image-to-image retrieval examples are shown in Figure~\ref{fig:i2i-retrieval-examples}.

\section{Detailed Ablation Results}
\label{app:detailed-ablation-results}

This subsection provides the full per-metric retrieval results for the ablation studies discussed in Section~\ref{sec:ablation_study}. Specifically, Table~\ref{tab:unlock_groups} reports results for partial fine-tuning of the image encoder, Tables~\ref{tab:clip_loss_lambda} and~\ref{tab:siglip_loss_lambda} report results for varying the hybrid loss weight $\lambda$, and Table~\ref{tab:combine_method_similarity_graph} reports results for different similarity graph combination strategies.

\section{Retrieval Results with Different Model Architectures}
\label{app:retrieval-results-with-different-model-architectures}

To verify whether SIGROT improves performance across different model architectures, an ablation study was conducted by alternating the image and text towers with different pretrained models. For all main experiments presented in Section~\ref{sec:experiments}, the text tower uses an SBERT~\cite{reimers-2019-sentence-bert} model fine-tuned on Vietnamese corpora\footnote{\url{https://huggingface.co/keepitreal/vietnamese-sbert}} and the image tower uses ViT-B/16 pretrained with DINOv3~\cite{simeoni2025dinov3}. In this ablation, additional text encoders are evaluated, including EmbeddingGemma-300M\footnote{\url{https://huggingface.co/google/embeddinggemma-300m}}~\cite{vera2025embeddinggemma} and BGE-M3\footnote{\url{https://huggingface.co/BAAI/bge-m3}}~\cite{chen2024bge}, both of which are top-performing small-size multilingual models on the MTEB leaderboard\footnote{\url{https://huggingface.co/spaces/mteb/leaderboard}}~\cite{muennighoff2022mteb,enevoldsen2025mmtebmassivemultilingualtext} with support for Vietnamese. For the image tower, ConvNeXt-base\footnote{\url{https://huggingface.co/timm/convnext_base.dinov3_lvd1689m}} pretrained with DINOv3 is also considered as an alternative architecture.

Detailed retrieval results are presented in Table~\ref{tab:retrieval-results-with-different-model-architectures}. Across all tested configurations, integrating SIGROT with SigLIP consistently improves retrieval performance compared to using SigLIP alone. These results demonstrate that the effectiveness of SIGROT is not limited to a specific architecture but generalizes across different encoder choices. Notably, the SBERT model fine-tuned specifically for Vietnamese exhibits strong robustness against larger multilingual text encoders such as EmbeddingGemma-300M and BGE-M3. While using BGE-M3 as the text tower yields the highest overall retrieval results, the improvement over SBERT is marginal despite being roughly 3$\times$ larger in model size. On the image encoder side, ConvNeXt-base performs behind ViT-B/16 across all configurations.

Furthermore, the alignment score increases and the modality gap decreases consistently when SIGROT is applied, regardless of the model architecture, as shown in Table~\ref{tab:alignment-modality-gap-different-architectures}. This indicates that SIGROT effectively bridges the modality gap by aligning image and text representations more closely in the shared embedding space, confirming its generalizability across diverse model architectures.

\begin{table*}[tbp]
\centering
\small
\caption{Effect of partial fine-tuning of the image encoder on retrieval performance using ViCLIP-OT.
The row with 0 represents a fully frozen image encoder, while $k$ indicates unfreezing the last $k$ groups (14 corresponds to all layers).}
\label{tab:unlock_groups}
\resizebox{.9\textwidth}{!}{%
\begin{tabular}{c ccc ccc | c}
\toprule
\multirow{2}{*}{\makecell{\texttt{\#} Last Unfrozen\\ Groups}}
& \multicolumn{3}{c}{Text $\rightarrow$ Image}
& \multicolumn{3}{c|}{Image $\rightarrow$ Text}
& \multirow{2}{*}{\makebox[1.05cm]{Avg.}} \\
\cmidrule(lr){2-4} \cmidrule(lr){5-7}
& \makebox[1.05cm]{R@1}
& \makebox[1.05cm]{R@5}
& \makebox[1.05cm]{R@10}
& \makebox[1.05cm]{R@1}
& \makebox[1.05cm]{R@5}
& \makebox[1.05cm]{R@10}
& \\
\midrule
0 & 28.06 & 55.94 & 67.48 & 44.36 & 73.45 & 82.79 & 58.68 \\
1 & 28.21 & 56.12 & 67.64 & 43.76 & 74.14 & 83.89 & 58.96 \\
2 & 35.91 & 63.63 & 74.17 & 52.86 & 80.11 & 87.62 & 65.72 \\
3 & 39.01 & 66.04 & 75.91 & 55.49 & 81.20 & 89.16 & 67.80 \\
5 & 40.96 & 67.96 & 77.01 & 57.28 & 82.35 & 90.05 & 69.27 \\
7 & 40.65 & 68.41 & 77.79 & 57.28 & 82.00 & 90.15 & 69.38 \\
9 & 41.33 & 68.28 & 77.49 & 57.24 & 82.35 & 90.90 & \underline{69.60} \\
11 & 40.98 & 68.18 & 77.61 & 56.49 & 82.45 & 90.01 & 69.29 \\
13 & 41.47 & 68.26 & 77.77 & 57.09 & 83.29 & 89.86 & \textbf{69.62} \\
14 & 41.01 & 67.67 & 77.39 & 56.29 & 82.05 & 89.81 & 69.04 \\
\bottomrule
\end{tabular}
}%
\end{table*}

\begin{table*}[btp]
\centering
\caption{Effect of the CLIP loss weight $\lambda$ on retrieval performance using the proposed hybrid loss ($\lambda=0$ corresponds to SIGROT only). The highest scores are highlighted in bold, and the second-highest scores are underlined.}
\label{tab:clip_loss_lambda}
\resizebox{.775\textwidth}{!}{%
\begin{tabular}{c ccc ccc | c}
\toprule
\multirow{2}{*}{\makebox[1.05cm]{$\lambda$}} & \multicolumn{3}{c}{Text $\rightarrow$ Image} & \multicolumn{3}{c|}{Image $\rightarrow$ Text} & \multirow{2}{*}{\makebox[1.05cm]{Avg.}} \\
\cmidrule(lr){2-4} \cmidrule(lr){5-7}
& \makebox[1.05cm]{R@1} & \makebox[1.05cm]{R@5} & \makebox[1.05cm]{R@10} & \makebox[1.05cm]{R@1} & \makebox[1.05cm]{R@5} & \makebox[1.05cm]{R@10} & \\
\midrule
0.0 & 38.48 & 64.47 & 74.59 & 35.90 & 58.83 & 68.47 & 56.79 \\
0.1 & 41.01 & 67.67 & 77.39 & 56.29 & 82.05 & 89.81 & \underline{69.04} \\
0.2 & 39.94 & 67.62 & 77.24 & 56.74 & 82.89 & 90.75 & \textbf{69.20} \\
0.3 & 38.38 & 66.67 & 76.49 & 55.10 & 82.30 & 89.71 & 68.11 \\
0.5 & 37.40 & 65.88 & 75.76 & 53.26 & 81.15 & 88.96 & 67.07 \\
1.0 & 36.27 & 65.03 & 75.13 & 52.26 & 79.61 & 87.52 & 65.97 \\
2.0 & 35.22 & 63.85 & 74.25 & 50.82 & 77.97 & 86.97 & 64.85 \\
\bottomrule
\end{tabular}
}%
\end{table*}
\begin{table*}[htbp]
\centering
\caption{Effect of the SigLIP loss weight $\lambda$ on retrieval performance using the proposed hybrid loss ($\lambda=0$ corresponds to SIGROT only). The highest scores are highlighted in bold, and the second-highest scores are underlined.}
\label{tab:siglip_loss_lambda}
\resizebox{.775\textwidth}{!}{%
\begin{tabular}{c ccc ccc | c}
\toprule
\multirow{2}{*}{\makebox[1.05cm]{$\lambda$}} & \multicolumn{3}{c}{Text $\rightarrow$ Image} & \multicolumn{3}{c|}{Image $\rightarrow$ Text} & \multirow{2}{*}{\makebox[1.05cm]{Avg.}} \\
\cmidrule(lr){2-4} \cmidrule(lr){5-7}
& \makebox[1.05cm]{R@1} & \makebox[1.05cm]{R@5} & \makebox[1.05cm]{R@10} & \makebox[1.05cm]{R@1} & \makebox[1.05cm]{R@5} & \makebox[1.05cm]{R@10} & \\
\midrule
0.0 & 38.48 & 64.47 & 74.59 & 35.90 & 58.83 & 68.47 & 56.79 \\
0.1 & 41.51 & 68.76 & 77.91 & 59.08 & 85.33 & 91.99 & \textbf{70.76} \\
0.15 & 40.91 & 68.46 & 77.88 & 58.53 & 84.49 & 91.35 & \underline{70.27} \\
0.2 & 40.69 & 68.27 & 77.50 & 58.48 & 84.24 & 91.00 & 70.03 \\
0.3 & 40.20 & 67.84 & 77.10 & 57.88 & 83.49 & 90.20 & 69.45 \\
0.5 & 39.35 & 67.31 & 76.68 & 57.09 & 82.84 & 89.96 & 68.87 \\
\bottomrule
\end{tabular}
}%
\end{table*}

\begin{table*}[hbtp]
\centering
\small
\caption{Retrieval results with different model architectures. Results are reported on the UIT-OpenViIC test set. EG-300M denotes EmbeddingGemma-300M. ViT-B/16 and ConvNeXt-base are pretrained with the DINOv3 framework. Integrating SIGROT consistently improves retrieval performance across all model architectures.}
\label{tab:retrieval-results-with-different-model-architectures}
\resizebox{\textwidth}{!}{%
\begin{tabular}{l l c l ccc ccc | c}
\toprule
\multirow{2}{*}{\makecell{Text tower}}
& \multirow{2}{*}{\makecell{Image tower}}
& \multirow{2}{*}{\makecell{\texttt{\#} Params}}
& \multirow{2}{*}{\makecell{Method}}
& \multicolumn{3}{c}{Text $\rightarrow$ Image}
& \multicolumn{3}{c|}{Image $\rightarrow$ Text}
& \multirow{2}{*}{\makebox[1.05cm]{Avg.}} \\
\cmidrule(lr){5-7} \cmidrule(lr){8-10}
& & & & \makebox[1.05cm]{R@1}
& \makebox[1.05cm]{R@5}
& \makebox[1.05cm]{R@10}
& \makebox[1.05cm]{R@1}
& \makebox[1.05cm]{R@5}
& \makebox[1.05cm]{R@10}
& \\
\midrule
\multirow{2}{*}{SBERT} & \multirow{2}{*}{ViT-B/16} & \multirow{2}{*}{221M} & SigLIP & 34.75 & 63.01 & 72.96 & 50.10 & 79.78 & 88.04 & 64.77 \\
& & & SigLIP-SIGROT & 39.19 & 66.71 & 76.04 & 57.21 & 83.83 & 90.79 & 68.96 \\
\midrule
\multirow{2}{*}{EG-300M} & \multirow{2}{*}{ViT-B/16} & \multirow{2}{*}{394M} & SigLIP & 34.02 & 60.89 & 71.33 & 51.60 & 79.88 & 87.89 & 64.27 \\
& & & SigLIP-SIGROT & 39.63 & 65.93 & 75.07 & 57.61 & 83.58 & 90.04 & 68.64 \\
\midrule
\multirow{2}{*}{BGE-M3} & \multirow{2}{*}{ViT-B/16} & \multirow{2}{*}{655M} & SigLIP & 34.88 & 62.41 & 72.70 & 52.30 & 81.03 & 89.14 & 65.41 \\
& & & SigLIP-SIGROT & 39.77 & 66.91 & 76.65 & 58.06 & 85.04 & 91.29 & 69.62 \\
\midrule
\multirow{2}{*}{SBERT} & \multirow{2}{*}{ConvNeXt-base} & \multirow{2}{*}{223M} & SigLIP & 32.83 & 60.96 & 71.46 & 48.10 & 77.33 & 86.19 & 62.81 \\
& & & SigLIP-SIGROT & 36.44 & 63.81 & 73.17 & 51.45 & 80.48 & 88.19 & 65.59 \\
\bottomrule
\end{tabular}
}%
\end{table*}

\begin{table*}[htbp]
\centering
\caption{Effect of different similarity graph combination strategies on retrieval performance. The cross-modality strategy achieves the highest average R@K for both ViCLIP-OT and ViSigLIP-OT, while the caption-only strategy yields the lowest performance.}
\label{tab:combine_method_similarity_graph}
\resizebox{\textwidth}{!}{%
\begin{tabular}{l ccc ccc | c}
\toprule
\multirow{2}{*}{Combination strategy} & \multicolumn{3}{c}{Text $\rightarrow$ Image} & \multicolumn{3}{c|}{Image $\rightarrow$ Text} & \multirow{2}{*}{\makebox[1.05cm]{Avg.}} \\
\cmidrule(lr){2-4} \cmidrule(lr){5-7}
& \makebox[1.05cm]{R@1} & \makebox[1.05cm]{R@5} & \makebox[1.05cm]{R@10} & \makebox[1.05cm]{R@1} & \makebox[1.05cm]{R@5} & \makebox[1.05cm]{R@10} & \\
\midrule
\multicolumn{7}{c}{\textit{ViCLIP-OT}} \\
\midrule
Arithmetic mean (Eq. \ref{eq:arithmetic_mean_similarity_graph}) & \underline{38.77} & \underline{66.09} & \underline{76.21} & \underline{54.85} & \underline{81.65} & \textbf{90.45} & \underline{68.00} \\
Harmonic mean (Eq. \ref{eq:harmonic_mean_similarity_graph}) & 37.02 & 64.13 & 74.80 & 49.78 & 80.01 & 88.21 & 65.66 \\
Caption only~\cite{sobal2024mathbbxsamplecontrastivelossimproving} & 33.29 & 59.24 & 69.26 & 42.37 & 70.01 & 80.26 & 59.07 \\
Image only & 32.18 & 60.08 & 70.94 & 47.79 & 76.83 & 86.28 & 62.35 \\
Cross modality (Sec. \ref{subsec:sigrot}) & \textbf{41.01} & \textbf{67.67} & \textbf{77.39} & \textbf{56.29} & \textbf{82.05} & \underline{89.81} & \textbf{69.04} \\
\midrule
\multicolumn{7}{c}{\textit{ViSigLIP-OT}} \\
\midrule
Arithmetic mean (Eq. \ref{eq:arithmetic_mean_similarity_graph}) & 39.53 & 66.83 & 76.50 & 57.28 & 82.05 & 89.66 & 68.64 \\
Harmonic mean (Eq. \ref{eq:harmonic_mean_similarity_graph}) & \underline{39.99} & \underline{67.19} & \underline{76.55} & \underline{57.43} & \underline{82.89} & \underline{90.30} & \underline{69.06} \\
Caption only~\cite{sobal2024mathbbxsamplecontrastivelossimproving} & 31.28 & 58.25 & 68.42 & 46.64 & 75.19 & 84.19 & 60.66 \\
Image only & 38.56 & 66.02 & 75.22 & 53.95 & 81.45 & 89.31 & 67.42 \\
Cross modality (Sec. \ref{subsec:sigrot}) & \textbf{41.51} & \textbf{68.76} & \textbf{77.91 }& \textbf{59.08} & \textbf{85.33} & \textbf{91.99} & \textbf{70.76} \\
\bottomrule
\end{tabular}
}%
\end{table*}

\begin{table*}[hbtp]
\centering
\small
\fontsize{10pt}{10pt}\selectfont
\caption{Alignment score and modality gap with different model architectures. $A$ denotes the alignment score (higher is better), and $\| \Delta_{\text{gap}} \|$ denotes the modality gap (lower is better).}
\label{tab:alignment-modality-gap-different-architectures}
\resizebox{.8\textwidth}{!}{%
\begin{tabular}{l l l c c}
\toprule
\makecell{Text tower}
& \makecell{Image tower}
& \makecell{Method}
& \makecell{A $\uparrow$}
& \makecell{$\| \Delta_{\text{gap}} \| \downarrow$} \\
\midrule
\multirow{2}{*}{SBERT} & \multirow{2}{*}{ViT-B/16} & SigLIP &  0.3637 & 0.5843 \\
& & SigLIP-SIGROT & 0.3928 & 0.3177 \\
\midrule
\multirow{2}{*}{EG-300M} & \multirow{2}{*}{ViT-B/16} & SigLIP & 0.3460 & 0.5234 \\
& & SigLIP-SIGROT & 0.3543 & 0.4066 \\
\midrule
\multirow{2}{*}{BGE-M3} & \multirow{2}{*}{ViT-B/16} & SigLIP & 0.3866 & 1.3584 \\
& & SigLIP-SIGROT & 0.4007 & 0.8405 \\
\midrule
\multirow{2}{*}{SBERT} & \multirow{2}{*}{ConvNeXt-base} & SigLIP & 0.3473 & 0.5835 \\
& & SigLIP-SIGROT & 0.3742 & 0.3340 \\
\bottomrule
\end{tabular}
}%
\end{table*}

\end{document}